\documentclass[12pt]{article}
\usepackage{float}
\usepackage{graphicx}
\usepackage[printonlyused]{acronym}
\usepackage{hyperref}
\usepackage{xcolor}
\usepackage[margin=0.8in]{geometry}
\usepackage[preprint]{neurips_2022}
\usepackage{amsmath}
\usepackage{multicol}
\usepackage{bm}
\usepackage[]{graphicx}
\usepackage{subfig}
\usepackage{color}
\usepackage{amssymb, epsfig}
\usepackage{url}
\usepackage{hyperref}
\usepackage{titlesec}
\usepackage{bbm}

\usepackage[utf8]{inputenc}
\usepackage[style=numeric, citestyle=numeric-comp]{biblatex}
\DefineBibliographyStrings{english}{
  in = {} 
}

\DeclareFieldFormat[article]{title}{#1}
\DeclareFieldFormat[book]{title}{#1}
\DeclareFieldFormat[inbook]{title}{#1}
\DeclareFieldFormat[incollection]{title}{#1}
\DeclareFieldFormat[inproceedings]{title}{#1}
\DeclareFieldFormat[thesis]{title}{#1}
\DeclareFieldFormat[misc]{title}{#1} 
\DefineBibliographyStrings{english}{
  in = {} 
}

\DeclareFieldFormat{labelnumber}{[#1]} 

\DeclareFieldFormat{labelnumber}{#1}
\DeclareFieldFormat{pages}{#1} 
\DeclareFieldFormat[article]{volume}{\textbf{#1}}

\title{Improving physics-informed neural network extrapolation via transfer learning and adaptive activation functions}

\author{%
A. Papastathopoulos-Katsaros$^{1,2,\dagger,*}$ \\
\texttt{athanasios.papastathopoulos-katsaros@bcm.edu} \\
\And
Alexandra Stavrianidi$^{3,\dagger,*}$ \\
\texttt{alexst@stanford.edu} \\
\And
Zhandong Liu$^{1,2}$ \\
\texttt{zhandong.liu@bcm.edu} \\
\\
$^{1}$Department of Pediatrics, Baylor College of Medicine, Houston TX 77030, USA \\
$^{2}$Data Science Center, Jan and Dan Duncan Neurological Research Institute, Texas Children's Hospital, Houston TX 77030, USA \\
$^{3}$Department of Mathematics, Stanford University, Stanford CA 94305, USA \\
\\
$^{\dagger}$These authors contributed equally \\
$^{*}$Corresponding authors
}

\begin{document}

\date{}
\maketitle

\begin{abstract}
Physics-Informed Neural Networks (PINNs) are deep learning models that incorporate the governing physical laws of a system into the learning process, making them well-suited for solving complex scientific and engineering problems. Recently, PINNs have gained widespread attention as a powerful framework for combining physical principles with data-driven modeling to improve prediction accuracy. Despite their successes, however, PINNs often exhibit poor extrapolation performance outside the training domain and are highly sensitive to the choice of activation functions (AFs). In this paper, we introduce a transfer learning (TL) method to improve the extrapolation capability of PINNs. Our approach applies transfer learning (TL) within an extended training domain, using only a small number of carefully selected collocation points. Additionally, we propose an adaptive AF that takes the form of a linear combination of standard AFs, which improves both the robustness and accuracy of the model. Through a series of experiments, we demonstrate that our method achieves an average of $ 40 \%$ reduction in relative $L_2$ error and an average of $ 50 \%$ reduction in mean absolute error in the extrapolation domain, all without a significant increase in computational cost. The code is available at \url{https://github.com/LiuzLab/PINN-extrapolation}
\end{abstract}

\section{Introduction}
Physics-Informed Neural Networks (PINNs), introduced by Raissi et al. \cite{raissi2019physics}, are neural networks trained not only on data but also constrained by physical laws, expressed as partial differential equations (PDEs). By embedding physics as prior knowledge, PINNs restrict the space of admissible solutions and reduce overfitting. PINNs have been successful in various use cases, in fields such as fluid mechanics \cite{fluidmechanicsreview}, medicine \cite{MRIpaper,bloodpressure}, and earth science \cite{earthsciences}. In this work, we focus on using PINNs as solvers for nonlinear PDEs. 

We consider a nonlinear PDE of the form
$$
u_{t}+\mathcal{N}(u)=0, x \in D, t \in[0, T]
$$

\noindent where $u(t, x)$ denotes the latent solution, 
$\mathcal{N}$ denotes a nonlinear differential operator, $D$ is the spatial domain and $T$ is the final time. To ensure a well-posed problem, an initial condition is prescribed for $u$ at $t=0$, along with boundary conditions on $\partial{D}$. The mean squared error loss function of the neural network is given by
\begin{equation} \nonumber 
\begin{split}
&M S E= \frac{w_{u}}{N_{u}} \sum_{i=1}^{N_{u}}\left|u\left(t_{u}^{i}, x_{u}^{i}\right)-u^{i}\right|^{2}+  \frac{w_{f}}{N_{f}} \sum_{i=1}^{N_{f}}\left|u_{t}\left( t_{f}^{i}, x_{f}^{i} \right)+\mathcal{N}(u)\left(t_{f}^{i}, x_{f}^{i}\right) \right|^{2} \\ 
\end{split}
\end{equation}
where $\left\{t_{u}^{i}, x_{u}^{i}, u^{i}\right\}_{i=1}^{N_{u}}$ denote the initial and boundary data prescribed to the problem and $\left\{t_{f}^{i}, x_{f}^{i}\right\}_{i=1}^{N_{f}}$ denote the collocation points in the domain where the PDE is enforced. The weights $w_{u}$ and $w_{f}$ reflect our trust in the data and the PDE, respectively. 

Although PINNs require substantially less computational resources than finite elements methods \cite{periodicactivationsolutetransport,MRIpaper}, are mesh-free, and easily programmed, they perform poorly in extrapolation tasks \cite{dpm,harvard} which assess their performance on test data $\left\{\left(t_{\text {test }}^{i}, x_{\text {test }}^{i}\right)\right\} \subset \left(T_{\text {train }}, T\right] \times D $, where $T>T_{\text {train }}$. In this paper, we propose a transfer-learning method that reduces the $L_2$ extrapolation error by $ 40 \%$ on average and the mean absolute extrapolation error by $ 50 \%$ on average across several benchmarks, including the Allen–Cahn (AC), Korteweg–de Vries (KdV), and Burgers equations.

\section{Related Work}

The original PINN architecture \cite{raissi2019physics} uses the $\tanh$ activation function (AF) in all the layers. However, the performance of PINNs is strongly influenced by the choice of AF \cite{locallyadaptiveactivations,abbasi,abupinn,failperiodicfunctions}, and selecting AF that reflect the functional form of the solution has been shown to improve performance.  
In \cite{abbasi}, the authors use components of the PDE problem at hand (e.g., the PDE coefficients, the initial condition, analytical solutions in lower dimensions), while in \cite{failperiodicfunctions}, the authors propose using $x+\sin^{2}(x)$ to learn periodic functions. In \cite{locallyadaptiveactivations} and \cite{abupinn}, the authors introduce adaptive approaches. Following the ABU-PINN (Adaptive Blending Unit) method \cite{abupinn}, one can take a linear combination of standard AFs (GELU, ELU, sigmoid, tanh, sin, exp, Softplus, and Swish) with adaptive weights and adaptive slopes as the AF $f$: 

\begin{equation} \nonumber 
f(x)=\sum_{i=1}^{N} G\left(\alpha_{i}\right) \sigma_{i}\left(\beta_{i} x\right) 
\end{equation}
\noindent where $\sigma_{i}(\cdot)$ , $\alpha_{i}$ , $\beta_{i}$ denote the candidate AF, the learnable weight parameters, and the learnable scaling factors (adaptive slopes) respectively. The gate function $G(\cdot)$ is softmax,   $G\left(\alpha_{i}\right)=\exp \left(\alpha_{i}\right) / \sum_{j=1}^{N} \exp \left(\alpha_{j}\right)$, so the search space is the convex hull of the set of candidate AF. In the rest of the paper, we will refer to these as ABU-PINN combinations. 

The performance of the PINN also depends strongly on the selection of collocation points and the choice of the weights $w_u$ and $w_f$. As demonstrated in prior works \cite{allencahnadaptive,residualadaptivesampling}, adaptively sampling collocation points from regions that exhibit large PDE residuals during training can substantially improve accuracy.

Different PDEs have different generalization capabilities \cite{extrapolationdrivennetwork}, as equations with a higher time derivative $|u_{t}|$ exhibit more variance in the time domain. For some PDEs, the extrapolation error increases exponentially fast as one moves away from the training domain \cite{harvard}. In \cite{dpm}, the authors propose a method that adjusts the gradient descent vector update to always point in a direction that reduces the PDE loss error during training. However, this is ineffective when the initial and boundary conditions are imposed as hard constraints to accelerate training \cite{incorporateconditions}, as the loss function only includes the PDE residual.

Motivated by these considerations, we incorporate the initial and boundary conditions as hard constraints and propose a transfer-learning (TL) method to deal with the extrapolation issue, in which the AF selection plays a
crucial role. By retraining only the final layer of the network on a small set of carefully selected collocation points in the validation region and learning an optimal linear combination of AF, we improve the extrapolation performance substantially for several benchmarks on the Allen-Cahn (AC), Korteweg–de Vries (KdV), and Burgers equations. We achieve an average reduction of $ 40 \%$ for the $L_2$ and of $ 50 \% $ for the mean absolute extrapolation error.

By fine-tuning only the parameters of the final layer and those of the adaptive AF, we achieve this error reduction with a minimal computational cost (see Table \ref{trainingtimes}). Fine-tuning only the final layer's parameters is a common practice in transfer learning \cite{desai_one-shot_2021}, and this approach is further supported by a layer-wise feature importance analysis we conducted for the AC equation (see Figure S$\ref{grad_norms}$). 

Prior transfer learning approaches for PINNs have primarily focused on parameter tuning and data reuse to achieve generalization across different physical systems. In these works, transfer learning is applied across tasks, from one PDE to another, to improve performance in forward problems \cite{desai_one-shot_2021,kapoor_transfer_2024,prantikos_physics-informed_2023,mustajab_physics-informed_2024,shima_modeling_2024,jalili_transfer_2024}, and inverse problems \cite{zhu_transfer_2024}.
In contrast, our method focuses on enhancing extrapolation performance for the same PDE in regions beyond the training domain. We also use a fundamentally different strategy that combines adaptive activation function design with localized retraining of only the final layer based on newly sampled high-loss collocation points from an intermediate region.

\section{Method Description}
We split our temporal domain into three segments: the training range, the validation range, and the testing range, i.e.  $\left[0, T_{\text {train }}\right],\left(T_{\text {train }}, T_{\text {val }}\right]$, and $\left(T_{\text {val }}, T_{\text {test }}\right]$, where $T=T_{\text {test }}>T_{\text {val }}>T_{\text {train }}>0$.  We use $T_{\text {train }}=\frac{T}{2}, T_{\text {val }}=\frac{4 T}{5}$, and $T_{\text {test }}=T$. The final $20 \%$ of the temporal domain is reserved for extrapolation, as in \cite{dpm} and other time-series prediction studies.
\par To enforce the initial and boundary conditions of the PDE as hard constraints, we perform a change of variables to the governing equations. This ensures that the constraints are satisfied exactly and eliminates the need to balance the data loss against the PDE loss using fine-tuned weights $w_u$ and $w_f$.

\vspace{1em}
\noindent\textbf{Initial Training Phase}
\vspace{0.5em}

\begin{enumerate}
\item{\noindent We begin by training the neural network on the domain $[0, T_{train}]$. The training is monitored using the  $L_2$ error on the validation interval $[T_{train}, T_{val}]$, and early stopping is applied to prevent overfitting.
}
\item{ \noindent After early stopping, we assess the extrapolation performance by computing the $L_2$ error on the test interval $(T_{\text{val}}, T_{\text{test}}]$.
}
\item{ \noindent To investigate how different AF impact extrapolation, we fix the AF $\tanh$ in all the layers except the final one, where we experiment with different alternatives.
}
\end{enumerate}

\vspace{1em}
\noindent\textbf{Transfer Learning (TL) Phase}
\vspace{0.5em}

\begin{enumerate}
\item{ \noindent  After the initial training phase is complete, we freeze the weights of all the layers except the final one, which is a typical practice in TL.
}

\item{ \noindent We uniformly sample a relatively large number of collocation points from the interval $[0, T_{\text{val}})$ and calculate their respective loss in a forward pass of the network. We then select a very small number of those collocation points: the ones that led to the highest loss.}

\item{ \noindent
We retrain the network on those few collocation points, using the Adam optimizer, updating only the final layer's weights and the parameters of its AF. This helps refine the model’s predictive ability in the extrapolation domain without disrupting earlier learned dynamics. We use $L2$ regularization, which penalizes large deviations in the retrained weights by adding a term proportional to the square of their magnitude. Among several techniques to avoid catastrophic forgetting (including basic fine-tuning and Elastic Weight Consolidation \cite{catastrophic_forgetting}), $L2$ regularization yielded the best performance, so we adopted it for all our experiments (see supplementary material section A).}
\end{enumerate}

\section{Experimental Setup}
\subsection{Hyperparameter selection}
We applied our method to the Allen–Cahn (AC), Korteweg–de Vries (KdV), and Burgers equations using neural networks composed of six layers with 32 neurons each. We use the L-BFGS optimizer \cite{byrd_limited_1995}, which ensures rapid convergence, paired with early-stopping. The network parameters are initialized using the Xavier scheme. For all equations, we use $8k$ uniformly sampled collocation points. For the KdV equation specifically, we also include $200$ boundary points.
\par For the TL phase, we switch to the Adam optimizer for an additional 120 to 150 epochs, using a learning rate of \texttt{1e-3} to \texttt{1e-2}. A total of 80 collocation points are selected for training in the transfer phase out of $4k$ uniformly sampled collocation points.

\subsection{Sampling strategy}
We experimented with varying the proportion of collocation points sampled in the training interval $[0, 0.5]$ and in the validation interval $(0.5, 0.8]$. Ultimately, we found that sampling exclusively from the validation interval produced the best results. Additionally, the existing mechanisms in our approach, including $L2$ regularization, the chosen number of training epochs, and the step size adjustments during TL, were sufficient to prevent catastrophic forgetting. The corresponding results are presented in Figures S\ref{coll_points_ac}, S\ref{coll_points_kdv}, S\ref{coll_points_burgers}.

\subsection{ Evaluation metrics}
We consider the $L_{2}$ relative error and the mean absolute error (MAE) to evaluate the accuracy of the PINN solutions, defined as follows:
$$
L_{2} \text{ error}=\frac{\sqrt{\sum_{i=1}^{N}\left|u_{\theta}\left(x_{i}, t_{i}\right)-u\left(x_{i}, t_{i}\right)\right|^{2}}}{\sqrt{\sum_{i=1}^{N}\left|u\left(x_{i}, t_{i}\right)\right|^{2}}} 
$$
$$
M A E =\frac{{\sum_{i=1}^{N}\left|u_{\theta}\left(x_{i}, t_{i}\right)-u\left(x_{i}, t_{i}\right)\right|}}{\sum_{i=1}^{N}\left|u\left(x_{i}, t_{i}\right)\right|}
$$
where $N$ is the number of test points, $u\left(x_{i}, t_{i}\right)$ is the reference solution at $\left(x_{i}, t_{i}\right)$, and $u_{\theta}\left(x_{i}, t_{i}\right)$ is the PINN solution at $\left(x_{i}, t_{i}\right)$. To measure the extrapolation error we take $\left\{\left(x_{\text {test }}^{i}, t_{\text {test }}^{i}\right)\right\}$, where $x_{\text {test }}^{i}$ is sampled at a uniform spatial mesh grid in the spatial domain and $t_{\text {test }}^{i}$ is on a uniform temporal grid in $\left(T_{\text {val }}, T_{\text {test }}\right]$. We use a temporal step size of $0.01$ for Burgers, $0.005$ for AC and KdV, and sample 400 collocation points at each time instant for AC, 600 for Burgers, and 500 for KdV.
\par To obtain the reference solution to the PDE, we used a finite difference method for spatial discretization combined with the implicit ODE solver ode15s in MATLAB. The second spatial derivative was approximated using a centered difference scheme, and the other terms were evaluated pointwise. To enforce the boundary conditions, the solution at the domain boundaries was explicitly set at every time step. The solver's relative and absolute tolerances were set to \texttt{1e-6} and \texttt{1e-8}, respectively, ensuring numerical stability and accuracy.

\section{Experimental results}
We benchmarked our method on three different equations: Allen-Cahn (AC), Korteweg–de Vries (KdV), and the viscous Burgers equation. Each experiment was repeated ten times, and we report the average results. In all examples, the training domain was set to $[0,0.5]$, the validation domain to $[0.5, 0.8]$ and the extrapolation domain to $[0.8,1]$. As mentioned previously, we evaluated the extrapolation performance of different AFs in the final layer while all other layers used tanh. The pool of candidate AFs for the final layer consists of:
\begin{description}
    \item [tanh:] The most commonly used AF for PINNs.
    \item [ABU-PINN from \cite{abupinn}:] A linear combination of different AF including GELU, ELU, sigmoid, tanh, sin, exp, Softplus, and Swish. We take adaptive linear combinations of at least three and up to six AF at a time. The use of softmax for weight normalization is maintained as in \cite{abupinn}. 
    \item  [\mathversion{bold}
$x + \sin^2(x)$ from \cite{failperiodicfunctions}:] It has been shown to improve performance, potentially due to its semi-periodic component and its monotonicity.
    \item [Linear combinations of tanh (lctanh):]
 $\sum_{i=1}^{n} w_i\tanh(a_i x + b_i)$, where the coefficients $a_i$, $b_i$, $w_i$ are optimized.
    \item [Linear combinations of sin (lcsin):] $\sum_{i=1}^{n} w_i\sin(a_i x + b_i)$, where the coefficients $a_i$, $b_i$, $w_i$ are optimized.
    \item [Linear combinations of\mathversion{bold}
$x + \sin^2(x)$:
\newline \mathversion{normal}]
$\sum_{i=1}^{n} ( a_i x + b_isin^2(c_ix+d_i))$, where the coefficients $a_i$, $b_i, c_i, d_i$ are optimized.
\end{description}

As noted in \cite{abupinn}, each PDE tends to have a preferred AF; for example, tanh performs best for the Allen–Cahn and viscous Burgers equations, while 
$x+ \sin^2 (x) $ is preferred for KdV. Building on this insight, we consider adaptive linear combinations of the preferred AF for each PDE (see supplementary material section D).

\subsection{Allen-Cahn (AC) equation}
The AC equation is given by
\nonumber
\begin{equation}
\begin{split}
&u_{t}-0.0001 u_{x x}+5 u^{3}-5 u=0, \text{ } x \in[-1,1],\text{ }  t \in[0,1], \\
&u(0,x)=x^{2} \cos (\pi x), u(-1, t)=u(1, t)=-1.
\end{split}
\end{equation}

\noindent We enforce the initial and boundary conditions as hard constraints by setting
$$
u(t, x)=x^{2} \cos (\pi x)+t\left(1-x^{2}\right) v(t, x). 
$$

\noindent Then $u$ automatically satisfies the initial and boundary conditions, and we only need to impose the PDE constraint on $v$ (see supplementary material section B). 

The best ABU-PINN was the combination of tanh, GELU, and sigmoid. For lctanh and lcsin, the combination of three AF was shown to give the best results. Our findings are presented in Table $\ref{table_ac}$. Overall, a linear combination of tanh provided the best extrapolation accuracy while introducing a similar number of trainable parameters as other linear combinations.
\par Regarding the TL results, we plot the collocation points with the highest PDE loss (the ones chosen for training) in Figure S\ref{coll_points_ac}. As expected, these points are predominantly concentrated towards the end of the temporal validation domain and around the two regions with the largest spatial discrepancies. We trained for 150 epochs and set the learning rate for the Adam optimizer to \texttt{5e-3}.
\par The performance of various AF with TL is summarized in Table \ref{table_ac}. Despite using only a small number of collocation points and training for a limited number of epochs, our results demonstrate substantial improvements. Specifically, we reduced the $L_2$ error from $0.20$ to $0.12$ compared to the standard PINN model and the MAE error from $0.11$ to $0.06$. A benchmark of our method against the state-of-the-art methods of References \cite{dpm,mcclenny_self-adaptive_2023,multinetwork} is presented in Table \ref{benchmark_ac}. The results show that our TL approach outperforms all other methods.
\par Figures S\ref{tl_comparison_ac_1}, S\ref{tanh_comparison_ac_1}, S\ref{tl_comparison_ac_2}, and S\ref{tanh_comparison_ac_2} allow for a visual comparison of the model performance with and without TL, as well as the performance of the tanh AF versus the best lctanh. The results are shown for two different timestamps, $t=0.82$ and $t=0.99$, which highlight the challenges of extrapolation due to the rapid spatial transitions that occur over time. As illustrated in the figures, it is evident that both TL and a well-chosen AF contribute significantly to the improvement of the extrapolation performance. Furthermore, errors tend to be larger at later timestamps ($t$), a finding that aligns with \cite{harvard}.
 
\begin{table}[ht] 
\begin{center}
\begin{tabular}{|l|ll|ll|}
\hline
                            & \multicolumn{2}{c|}{$L_2$}                   & \multicolumn{2}{c|}{MAE}                  \\ \hline
AF                          & \multicolumn{1}{l|}{Without TL} & With TL & \multicolumn{1}{l|}{Without TL} & With TL \\ \hline
tanh                        & \multicolumn{1}{l|}{0.20}       & 0.17& \multicolumn{1}{l|}{0.11}       & 0.09\\ \hline
$x + sin^2(x)$ & \multicolumn{1}{l|}{0.23}       & 0.19& \multicolumn{1}{l|}{0.14}       & 0.1\\ \hline
ABU-PINN                    & \multicolumn{1}{l|}{0.19}       & 0.14& \multicolumn{1}{l|}{0.11}       & 0.08    \\ \hline
lctanh                      & \multicolumn{1}{l|}{0.16}       & 0.12& \multicolumn{1}{l|}{0.09}       & 0.06\\ \hline
lcsin                       & \multicolumn{1}{l|}{0.18}       & 0.14& \multicolumn{1}{l|}{0.10}       & 0.07\\ \hline
\end{tabular}
\end{center}
 \caption{Comparative analysis of $L_2$ and MAE errors for various AF applied to the AC equation, demonstrating the effectiveness of TL and of carefully selected AF. Results for lctanh, lcsin, and ABU-PINN correspond to the optimal linear combinations described in the main text.}
\label{table_ac} 
\end{table}

\begin{table}[h!]
\begin{center}
\begin{tabular}{|l|l|}
\hline
Method               & $L_2$ extrapolation error \\ \hline
lctanh w/o TL (ours) & 0.16     \\ \hline
lctanh w/ TL (ours)  & 0.12     \\ \hline
SA-PINN \cite{mcclenny_self-adaptive_2023}     & 0.18     \\ \hline
w-s PINN \cite{multinetwork}    & 0.14     \\ \hline
DPM \cite{dpm}        & 0.18     \\ \hline
\end{tabular}
\end{center}
\caption{Comparison of our methods with state-of-the-art approaches applied to the AC equation.}
\label{benchmark_ac}
\end{table}

\subsection{ Korteweg–de Vries (KdV) equation}
The KdV equation is given by
\begin{equation}
\begin{split}
&u_{t}+u u_{x}+0.0025 u_{x x x}=0, \text{ } x \in[-1,1], \text{ } t \in[0,1], \\
&u(0, x)=\cos (\pi x), u(t,-1)=u(t, 1), u_{x}(t,-1)=u_{x}(t, 1).\\
\end{split}
\end{equation}
\noindent We enforce the initial condition as a hard constraint by setting 

$$
u(t, x)=\cos (\pi x)+t v(t, x) 
$$

\noindent Then $u$ automatically satisfies the initial condition, and we only need to impose the PDE constraint and the boundary conditions on $v$ (see supplementary material section B).

The best ABU-PINN was the combination of tanh, GELU, and sin. For lctanh and lcsin, the combination of three AFs yielded the best results, while the combination of two AFs achieved similar performance. Our findings are presented in Table $\ref{table_kdv}$. Overall, a linear combination of $x + \sin^2(x)$ with $n=2$ provided the best extrapolation accuracy while introducing a similar number of trainable parameters as other linear
combinations. As is the case for AC, we observe that the best adaptive linear combination is the linear combination of the single preferred AF, in this case, $x+\sin^2(x)$.
\par Regarding TL, we plot the collocation points with the highest PDE loss in Figure S\ref{coll_points_kdv}. We trained for 150 epochs and set the learning rate for the Adam optimizer to \texttt{5e-2}. The performance of the different AF with TL is summarized in Table~\ref{table_kdv}. Our method reduced the $L_2$ error from $0.15$ to $0.08$ compared to the standard PINN model and the MAE error from $0.12$ to $0.06$. A benchmark of our method against the state-of-the-art method of Reference \cite{multinetwork} can be found in Table \ref{benchmark_kdv}. The results show that our TL approach outperforms the other method. Figures S\ref{tl_comparison_kdv_1}, S\ref{tanh_comparison_kdv_1}, S\ref{tl_comparison_kdv_2}, and S\ref{tanh_comparison_kdv_2} present example calculations that compare model performance with and without TL, as well as the performance of the tanh AF versus the best lc$(x + sin^2(x))$.

\begin{table}[ht]
\begin{center}
\begin{tabular}{|l|ll|ll|}
\hline
                               & \multicolumn{2}{c|}{$L_2$}                   & \multicolumn{2}{c|}{MAE}                  \\ \hline
AF                             & \multicolumn{1}{l|}{Without TL} & With TL & \multicolumn{1}{l|}{Without TL} & With TL \\ \hline
tanh                           & \multicolumn{1}{l|}{0.15}       & 0.13    & \multicolumn{1}{l|}{0.12}       & 0.10    \\ \hline
$x + sin^2(x)$    & \multicolumn{1}{l|}{0.12}       & 0.09    & \multicolumn{1}{l|}{0.08}       & 0.07    \\ \hline
ABU-PINN                       & \multicolumn{1}{l|}{0.14}       & 0.11    & \multicolumn{1}{l|}{0.09}       & 0.08    \\ \hline
lctanh                         & \multicolumn{1}{l|}{0.14}       & 0.12    & \multicolumn{1}{l|}{0.11}       & 0.09    \\ \hline
lcsin                          & \multicolumn{1}{l|}{0.12}       & 0.09    & \multicolumn{1}{l|}{0.07}       & 0.07    \\ \hline
lc ($x + sin^2(x)$) & \multicolumn{1}{l|}{0.10}       & 0.08    & \multicolumn{1}{l|}{0.07}       & 0.06    \\ \hline
\end{tabular}
\end{center}
\caption{Comparative analysis of $L_2$ and MAE errors for various AF applied to the KdV equation, demonstrating the effectiveness of TL and of carefully selected AF. Results for lctanh, lcsin, lc($x + sin^2(x)$), and ABU-PINN correspond to the optimal linear combinations described in the main text.}
\label{table_kdv}
\end{table}

\begin{table}[h!]
\begin{center}
\begin{tabular}{|l|l|}
\hline
Method                & $L_2$ extrapolation error \\ \hline
lc ($x + sin^2(x)$) w/o TL (ours) & 0.10     \\ \hline
lc ($x + sin^2(x)$) w/ TL (ours)  & 0.08     \\ \hline
s-d PINN \cite{multinetwork}   & 0.14     \\ \hline
\end{tabular}
\end{center}
\caption{Comparison of our methods with state-of-the-art approaches applied to the KdV equation.}
\label{benchmark_kdv}
\end{table}
\subsection{ Viscous Burgers' equation}
The viscous Burgers' equation is given by
\begin{equation}
\begin{split}
&u_{t}+u u_{x}-\frac{0.01}{\pi} u_{x x}=0, \text{ }x \in[-1,1], \text{ } t \in[0,1], \\
&u(0,x)=-\sin (\pi x), u(t,-1)=u(t, 1)=0.\\
\end{split}
\end{equation}

\noindent We enforce the initial and boundary conditions as hard constraints by setting 
$$
u(t,x)=-\sin (\pi x) + t(1-x^2)v(t,x)
$$

\noindent Then u automatically satisfies the initial and boundary conditions, and we only need to impose
the PDE constraint on $v$. 

\noindent The best ABU-PINN was the combination of tanh, GELU, sigmoid, and sin. For lctanh and lcsin, the combination of three AFs yielded the best results. Our results are summarized in Table $\ref{table_burgers}$. The linear combination of tanh provides the best extrapolation accuracy, and we see again that the adaptive linear combination that performs best is the linear combination of the single preferred AF.
\par We plot the collocation points with the highest PDE loss in Figure S\ref{coll_points_burgers}. As before, these points are predominantly concentrated towards the end of the temporal validation domain and around the region with the largest spatial discrepancies. We trained for 150 epochs and set the learning rate for the Adam optimizer to \texttt{5e-2}.
\par The performance of the various AF with TL is summarized in Table \ref{table_burgers}. We reduced the $L_2$ error from $0.09$ to $0.06$ compared to the standard PINN model, and the MAE error from $0.07$ to $0.03$. A benchmark of our method against the state-of-the-art methods of References \cite{dpm,mcclenny_self-adaptive_2023} can be found in Table \ref{benchmark_burgers}. The results show that our approaches outperform all other methods. Figures S\ref{tl_comparison_burgers_1}, S\ref{tanh_comparison_burgers_1}, S\ref{tl_comparison_burgers_2}, S\ref{tanh_comparison_burgers_2} present example calculations that compare model performance with and
without TL, as well as the performance of the tanh AF versus the best lctanh.

\begin{table}[ht]
\begin{center}
\begin{tabular}{|l|ll|ll|}
\hline
                            & \multicolumn{2}{c|}{$L_2$}                   & \multicolumn{2}{c|}{MAE}                  \\ \hline
AF                          & \multicolumn{1}{l|}{Without TL} & With TL & \multicolumn{1}{l|}{Without TL} & With TL \\ \hline
tanh                        & \multicolumn{1}{l|}{0.09}       & 0.08    & \multicolumn{1}{l|}{0.07}       & 0.06    \\ \hline
$x + sin^2(x)$ & \multicolumn{1}{l|}{0.08}       & 0.08    & \multicolumn{1}{l|}{0.04}       & 0.03    \\ \hline
ABU-PINN                    & \multicolumn{1}{l|}{0.11}       & 0.09    & \multicolumn{1}{l|}{0.04}       & 0.03    \\ \hline
lctanh                      & \multicolumn{1}{l|}{0.07}       & 0.06    & \multicolumn{1}{l|}{0.04}       & 0.03    \\ \hline
lcsin                       & \multicolumn{1}{l|}{0.10}       & 0.07    & \multicolumn{1}{l|}{0.04}       & 0.04    \\ \hline
\end{tabular}
\end{center}
\caption{Comparative analysis of $L_2$ and MAE errors for various AF applied to the Burgers' equation, demonstrating the effectiveness of TL and of carefully selected AF. Results for lctanh, lcsin, and ABU-PINN correspond to the optimal linear combinations described in the main text.}
\label{table_burgers}
\end{table}
\begin{table}[h!]
\begin{center}
\begin{tabular}{|l|l|}
\hline
Method               & $L_2$ extrapolation error \\ \hline
lctanh w/o TL (ours) & 0.07     \\ \hline
lctanh w/ TL (ours)  & 0.06     \\ \hline
SA-PINN \cite{mcclenny_self-adaptive_2023}     & 0.08     \\ \hline
DPM \cite{dpm}    & 0.09     \\ \hline
\end{tabular}
\end{center}
\caption{Comparison of our methods with state-of-the-art approaches applied to the Burgers' equation.}
\label{benchmark_burgers}
\end{table}
\section{Conclusion}
In this work, we introduced a transfer learning (TL) method paired with adaptive activation functions (AF) to improve the extrapolation performance of physics-informed neural networks (PINNs). Through a series of numerical experiments, we demonstrated that our approach significantly outperforms the vanilla PINN while requiring minimal additional training cost and is also superior to three state-of-the-art methods. Our results highlight that both the choice of AF and the use of targeted TL within the validation region contribute to substantial gains. We evaluated our method on three partial differential equations (PDEs) and achieved an average of $ 40 \%$ $L_2$ error reduction and an average of $ 50 \% $ mean absolute error reduction in the extrapolation domain. 
\par We observed across all tested PDEs that the points with the highest PDE loss were found in the validation region, primarily near the domain's boundary. Selecting even a small number of such high-loss collocation points for the TL and implementing a regularization scheme to prevent catastrophic forgetting resulted in notable improvements in extrapolation performance. 
\par Additionally, we found that an adaptive linear combination of the PDE-specific preferred AF outperformed combinations of other AF, including multiple diverse AFs. These results suggest that the key to improved extrapolation lies in the adaptability introduced by learnable AF weights and that increasing the diversity of AFs in the combination (as in ABU-PINN) is suboptimal.

However, our method was evaluated only on 1D PDEs over regular domains,  and its scalability to more complex systems remains uncertain. In higher dimensions, collocation point selection becomes more critical, and enforcing boundary conditions as hard constraints on irregular domains may not be feasible. Applying our approach in such settings may require substantial adaptations and increased computational resources. Nonetheless, our method can still be applied in cases where real-world data is available, as our adaptive AF and TL approaches can still be implemented.

\subsubsection*{Acknowledgments} This work was supported by the National Institutes of Health (NIH) under grant number 5R01HG011795 and by the Cancer Prevention and Research Institute of Texas (CPRIT) under grant number RP240131. The authors also gratefully acknowledge the ICANN 2025 reviewers for their valuable feedback.

\printbibliography

@article{raissi2019physics,
  title={Physics-informed neural networks: A deep learning framework for solving forward and inverse problems involving nonlinear partial differential equations},
  author={M. Raissi and P. Perdikaris and G.E. Karniadakis},
  journal={J. Comput. Phys.},
  volume={378},
  pages={pp. 686-707},
  year={2019},
  publisher={Elsevier}
}

@article{ earthsciences,
author={P. Borate and J. Rivière and C. Marone and A.Mali and D. Kifer and P. Shokouhi},
title={Using a physics-informed neural network and fault zone acoustic monitoring to predict lab earthquakes},
journal={Nat. Commun.},
volume={14},
year={2023}
}

@article{multinetwork,
author={M. Penwarden and A.D.Jagtap and S.Zhe and G.Karniadakis and R.MKirby}, 
title={A unified scalable framework
for causal sweeping strategies for physics-informed neural networks (PINNs)
and their temporal decompositions},
journal={J. Comput. Phys.},
volume={493},
year={2023},
page={112464}
}

@article{extrapolationdrivennetwork,
title = {An extrapolation-driven network architecture for physics-informed deep learning},
journal = {Neural Netw.},
volume = {183},
pages = {pp. 106998},
year = {2025},
author = {Y. Wang and Y. Yao and Z. Gao}
}

@article{catastrophic_forgetting,
  title={Overcoming catastrophic forgetting in neural networks},
  author={J. Kirkpatrick and R. Pascanu and N. Rabinowitz and J. Veness and G. Desjardins,  and A.A. Rusu and K. Milan and J. Quan and T. Ramalho,  and A. Grabska-Barwinska and D. Hassabis, and C. Clopath and D. Kumaran and R. Hadsell},
  journal={Proc. Natl. Acad. Sci.},
  year={2016},
  volume={114},
  pages={pp. 3521-3526}
}

@article{byrd_limited_1995,
	title = {A Limited Memory Algorithm for Bound Constrained Optimization},
	journal = {SIAM J. Sci. Comput.},
	volume = {16},
	author = {R.H. Byrd and P. Lu and J. Nocedal and C. Zhu},
	year = {1995},
	pages = {pp. 1190--1208},
}

@article{failperiodicfunctions,
 author = {Z. Liu and H. Tilman and U. Masahito},
journal= {Adv. Neural Inf. Process. Syst.},
 page = {1583--1594},
 title = {Neural Networks Fail to Learn Periodic Functions and How to Fix It},
 volume = {33},
 year = {2020}
}

@misc{harvard, 
author={L. Fesser and L.D'Amico-Wong and R. Qiu}, 
title={Understanding and mitigating extrapolation failures in physics-informed neural networks},
note={\textit{ arXiv e-prints} 2306.09478 (2023) }
}

@article{allencahnadaptive,
title={Solving Allen-Cahn and Cahn-Hilliard equations using the adaptive physics informed neural networks},
author={C.L. Wight and J. Zhao},
journal={Commun. Comput. Phys.},
volume={29},
year={2021},
pages={ pp. 930–954}
}

@misc{abupinn,
author = {Wang, Honghui and Lu, Lu and Song, Shiji and Huang, Gao},
title = {Learning Specialized Activation Functions for Physics-informed Neural Networks},
note={ \textit{ arXiv e-prints} 2308.04073 (2023)
}
}

@article{residualadaptivesampling,
title = {A comprehensive study of non-adaptive and residual-based adaptive sampling for physics-informed neural networks},
journal = {Comput. Methods Appl. Mech. Eng.},
volume = {403},
pages = { pp. 115671},
year = {2023},
author = {C. Wu and M. Zhu and Q. Tan and Y. Kartha and L. Lu},
}

@article{mcclenny_self-adaptive_2023,
	title = {Self-adaptive physics-informed neural networks},
	volume = {474},
	journal = {Journal of Computational Physics},
	author = {L. D. McClenny and U.M. Braga-Neto},
	year = {2023},
	pages = {111722},
}

@Article{periodicactivationsolutetransport,
AUTHOR = {S.A. Faroughi and R.Soltanmohammadi and P.Datta and S.K.Mahjour and S.Faroughi},
TITLE = {Physics-Informed Neural Networks with Periodic Activation Functions for Solute Transport in Heterogeneous Porous Media},
JOURNAL = {Mathematics},
VOLUME = {12},
YEAR = {2024},
NUMBER = {1},
ARTICLE-NUMBER = {63},
}

@article{MRIpaper,
author = {B. Zapf and J. Haubner and M. Kuchta and G. Ringstad and P. Eide and K. A. Mardal},
year = {2022},
title = {Investigating molecular transport in the human brain from MRI with physics-informed neural networks},
volume = {12},
journal = {Sci. Rep.}
}

@article{locallyadaptiveactivations,
author = {A. Jagtap and K. Kawaguchi and G. Karniadakis},
year = {2020},
title = {Locally adaptive activation functions with slope recovery for deep and physics-informed neural networks},
journal = {Proc. R. Soc. A}
}

@article{incorporateconditions,
  author={T. Alkhalifah, and X. Huang},
 journal={2022 IEEE International Conference on Image Processing (ICIP)}, 
  title={Direct Imaging Using Physics Informed Neural Networks}, 
  year={2022},
  pages={2781-2785}}

@article{bloodpressure,
title = {Machine learning in cardiovascular flows modeling: Predicting arterial blood pressure from non-invasive {4D flow MRI} data using physics-informed neural networks},
journal = {Comput. Methods Appl. Mech. Eng.},
volume = {358},
pages = {pp. 112623},
year = {2020},
author = {G. Kissas and Y. Yang and E. Hwuang and W.R. Witschey and J. A. Detre and P. Perdikaris}
}

@article{fluidmechanicsreview,
  title={Physics-informed neural networks for fluid mechanics: a review},
  author={S. Cai and Z. Mao and Z. Wang and M. Yin and G. Karniadakis},
  journal={Acta Mech. Sin.},
  year={2021},
  volume={37},
  pages={ pp. 1727-1738}
}

@article{abbasi,
title = {Physical activation functions: An approach for more efficient induction of physics into physics-informed neural networks},
journal = {Neurocomputing},
volume={608},
year = {2024},
author = {J. Abbasi and P.Ø. Andersen}
}

@article{dpm,
title = {{DPM:A Novel Training Method for Physics-Informed Neural Networks in Extrapolation}},
author = {J. Kim and K. Lee and D. Lee and J. N. Park},
year = {2021},
journal = "35th AAAI Conference on Artificial Intelligence",
pages = "8146--8154"
}

@article{kapoor_transfer_2024,
	title = {Transfer learning for improved generalizability in causal physics-informed neural networks for beam simulations},
	volume = {133},
	journal = {Eng. Appl. Artif. Intell.},
	author = {T. Kapoor and H. Wang and A. Núñez and R. Dollevoet},
	year = {2024},
	pages = {pp. 108085},
}

@misc{desai_one-shot_2021,
	title = {One-{Shot} {Transfer} {Learning} of {Physics}-{Informed} {Neural} {Networks}},
	copyright = {arXiv.org perpetual, non-exclusive license},
	publisher = {arXiv},
	author = {S. Desai and M. Mattheakis and H. Joy and P. Protopapas and S. Roberts},
	note = {\textit{ arXiv e-prints} 2110.11286v2 (2021)},
}

@article{prantikos_physics-informed_2023,
	title = {Physics-informed neural network with transfer learning ({TL}-{PINN}) based on domain similarity measure for prediction of nuclear reactor transients},
	volume = {13},
	number = {1},
	journal = {Sci Rep},
	author = {K. Prantikos and S. Chatzidakis and L. H. Tsoukalas and A. Heifetz},
	year = {2023},
	pages = {pp. 16840},
}

@article{zhu_transfer_2024,
	title = {A transfer learning enhanced physics-informed neural network for parameter identification in soft materials},
	volume = {45},
	number = {10},
	journal = {Appl. Math. Mech.-Engl. Ed.},
	author = {J. Zhu and Y. Xue and Z. Liu},
	year = {2024},
	pages = {pp. 1685--1704},
}

@article{mustajab_physics-informed_2024,
	title = {Physics-{Informed} {Neural} {Networks} for {High}-{Frequency} and {Multi}-{Scale} {Problems} {Using} {Transfer} {Learning}},
	volume = {14},
	copyright = {https://creativecommons.org/licenses/by/4.0/},
	number = {8},
	journal = {Appl. Sci.},
	author = {A. H. Mustajab and H. Lyu and Z. Rizvi and F. Wuttke},
	year = {2024},
	pages = { pp. 3204},
}

@article{shima_modeling_2024,
	title = {Modeling unobserved geothermal structures using a physics-informed neural network with transfer learning of prior knowledge},
	volume = {12},
	number = {1},
	journal = {Geotherm Energy},
	author = {A. Shima and K. Ishitsuka and W. Lin and E. Bjarkason and A. Suzuki},
	year = {2024},
	pages = {pp. 38},
}

@article{jalili_transfer_2024,
	title = {Transfer learning through physics-informed neural networks for bubble growth in superheated liquid domains},
	volume = {232},
	journal = {Int. J. Heat Mass Transf.},
	author = {D. Jalili and M. Jadidi and A. Keshmiri and B. Chakraborty and A. Georgoulas and Y. Mahmoudi},
	year = {2024},
	pages = {pp. 125940},
}

\renewcommand{\tablename}{Table S}

\setcounter{table}{0}
\setcounter{section}{0}
\newpage
\section*{Supplementary materials}
\renewcommand{\thesection}{\Alph{section}}                     
\renewcommand{\thesubsection}{\thesection\arabic{subsection}} 
\setcounter{section}{0}
\setcounter{subsection}{0}
\section{Comparison of TL Methods }\label{compare_tl}
In this study, we examined three TL methods: vanilla TL, $L2$ regularization, and Elastic Weight Consolidation (EWC) \cite{catastrophic_forgetting}. We implemented all three methods by adjusting only the parameters of the final layer of the network while keeping all hyperparameters the same for a fair comparison.
\par In the vanilla model, the neural network was trained using standard backpropagation without any additional regularization terms. This approach serves as a baseline for comparing the effectiveness of advanced regularization techniques. 
\par $L2$ regularization modifies the loss function to prevent the model from overfitting to the new collocation points and avoid catastrophic forgetting, aiming to improve its generalizability. The modified loss function is given by:
\begin{equation*}
    \mathcal{L}(\theta) = \mathcal{L}_{\text{new}}(\theta) + \lambda \sum_{i} \theta_i^2
\end{equation*}
where:
\begin{itemize}
    \item $\mathcal{L}(\theta)$ is the total loss function,
    \item $\mathcal{L}_{\text{new}}(\theta)$ is the loss function for the new task, measuring the loss at the new collocation points,
    \item $\theta$ represents the parameters of the neural network,
    \item $\theta_i$ is the $i$-th parameter of the model,
    \item $\lambda$ is a regularization parameter that controls the trade-off between fitting the data well and keeping the model weights small to avoid overfitting. We experimented with multiple values and found $0.01$ to be optimal across all the PDEs we tested.
\end{itemize}
\par EWC is a method specifically designed to mitigate catastrophic forgetting during the TL process. To the best of our knowledge, it has not been tested on PINNs. It adds a penalty to the loss function, which is proportional to the square of the difference between the current and previous weights, scaled by the importance of the weights to previous tasks. Mathematically, it is given by:
\begin{equation*}
    \mathcal{L}(\theta) = \mathcal{L}_{\text{new}}(\theta) + \sum_{i} \frac{\lambda}{2} F_i (\theta_i - \theta_i^*)^2
\end{equation*}

where:
\begin{itemize}
    \item $\mathcal{L}(\theta)$ is the total loss function,
    \item $\mathcal{L}_{\text{new}}(\theta)$ is the loss function for the new task, measuring the loss at the new collocation points,
    \item $\theta$ represents the parameters of the neural network,
    \item $\theta_i$ is the current value of the $i$-th parameter,
    \item $\theta_i^*$ is the value of the $i$-th parameter after training on previous tasks,
    \item $F_i$ is the diagonal element of the Fisher Information Matrix for the $i$-th parameter, indicating its importance to previous tasks,
    \item $\lambda$ is a hyperparameter that controls the strength of the regularization. We experimented with multiple values and found $0.001$ to be optimal across all the PDEs we tested.
\end{itemize}
\par Table S1 summarizes the performance of each TL method on the AC equation, noting the increase of the $L2$ error in the training domain and the decrease of $L2$ error in the extrapolation domain (denoted by catastrophic forgetting rate and extrapolation error reduction, respectively). We used the ABU-PINN AF in the final layer, i.e. an adaptive linear combination of tanh, GELU, and sigmoid. The evaluation metrics consist of the relative $L_2$ error and the mean absolute error (MAE). Surprisingly, our findings reveal that EWC performs the worst, while the vanilla TL and the $L2$ regularization method exhibit similar results. Notably, $L2$ regularization presented an advantage, so we included it in all our experiments.

\begin{table}[ht]
\begin{center}
\begin{tabular}{|l|ll|ll|}
\hline
          & \multicolumn{2}{c|}{Catastrophic forgetting rate within [0.0,0.8]}     & \multicolumn{2}{c|}{Extrap. error reduction within [0.8,1.0]}   \\ \hline
TL method & \multicolumn{1}{l|}{L2 \%}  & MAE \% & \multicolumn{1}{l|}{L2 \%}  & MAE \% \\ \hline
Vanilla   & \multicolumn{1}{l|}{7.7}  & 38.7 & \multicolumn{1}{l|}{36.9} & 43.2 \\ \hline
L2 reg.    & \multicolumn{1}{l|}{5.2}  & 64.7 & \multicolumn{1}{l|}{40.1} & 43.8 \\ \hline
EWC       & \multicolumn{1}{l|}{15.6} & 68.3 & \multicolumn{1}{l|}{11.4} & 15.6 \\ \hline
\end{tabular}
\end{center}
\caption{Summary of catastrophic forgetting and extrapolation performance for the AC equation using the ABU-PINN AF across different TL methods.}
\label{tab:tl_methods}
\end{table}

\renewcommand{\figurename}{Figure S}

\setcounter{figure}{0}


\section{Experimental results}
\renewcommand{\thesubsubsection}{\thesubsection\arabic{subsubsection}}

\subsection{Allen-Cahn equation}

The AC equation is given by
\nonumber
\begin{equation}
\begin{split}
&u_{t}-0.0001 u_{x x}+5 u^{3}-5 u=0, \text{ } x \in[-1,1],\text{ }  t \in[0,1], \\
&u(0,x)=x^{2} \cos (\pi x), u(-1, t)=u(1, t)=-1.
\end{split}
\end{equation}

\noindent We enforce the initial and boundary conditions as hard constraints by setting
$$
u(t, x)=x^{2} \cos (\pi x)+t\left(1-x^{2}\right) v(t, x). 
$$

\noindent Then $u$ automatically satisfies the initial and boundary conditions, and we only need to impose the PDE constraint on $v$, which satisfies the equation

\begin{equation} \label{allencahneqv}
\begin{split}
&(1-x^2)v+t(1-x^2)v_{t} \\
&-0.0001\left[2 \cos ( \pi x) -4 \pi x \sin (\pi x)  -\pi^2 x^2\cos( \pi x) +t\left(-2v -4xv_{x}+(1-x^2)v_{xx}\right) \right] \\
&+5( x^{2} \cos (\pi x)+t\left(1-x^{2}\right) v(t, x) )^{3}-5 \left(x^{2} \cos (\pi x)+t\left(1-x^{2}\right) v(t, x)\right) =0 \\
\end{split}
\end{equation}

\begin{figure}[H]
    \centering
    \includegraphics[width=1.0\textwidth]{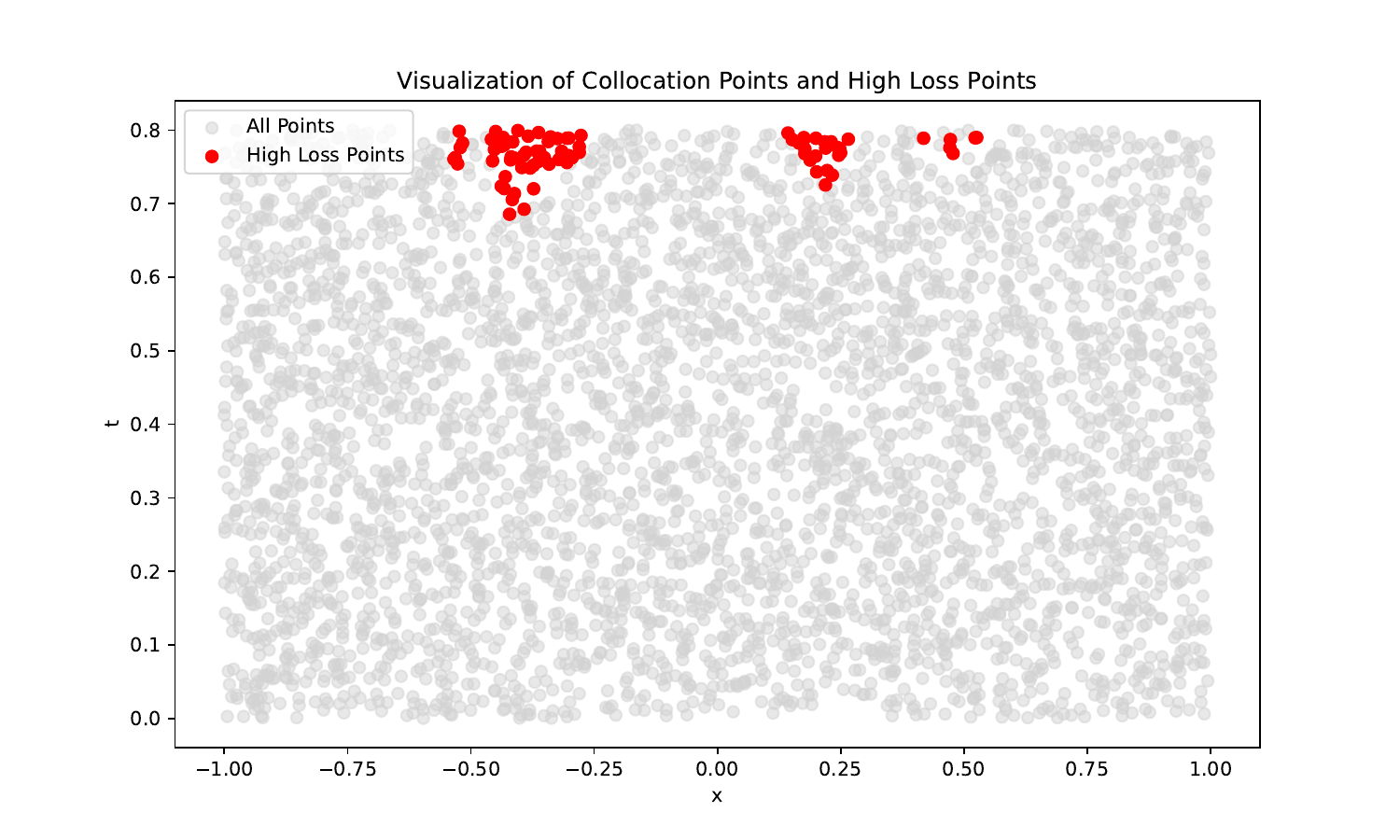}
    \caption{The collocation points with the highest PDE loss for the AC equation are concentrated near the end of the temporal validation domain and around the regions of largest discrepancy in the spatial domain.}
    \label{coll_points_ac}
\end{figure}

\begin{figure}[H]
    \centering
    \subfloat[Without TL]{\includegraphics[width=0.55\textwidth]{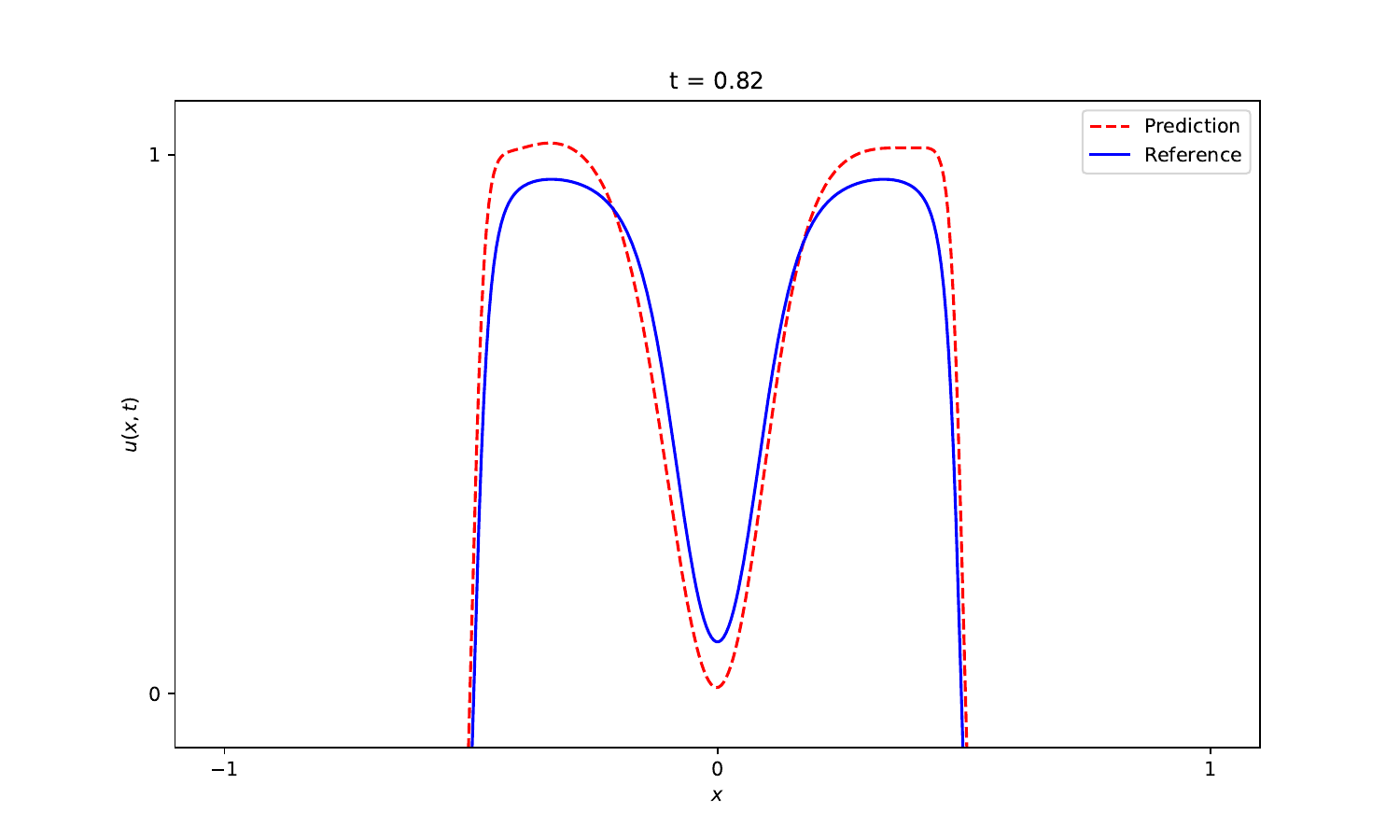}} 
    \subfloat[With TL]{\includegraphics[width=0.55\textwidth]{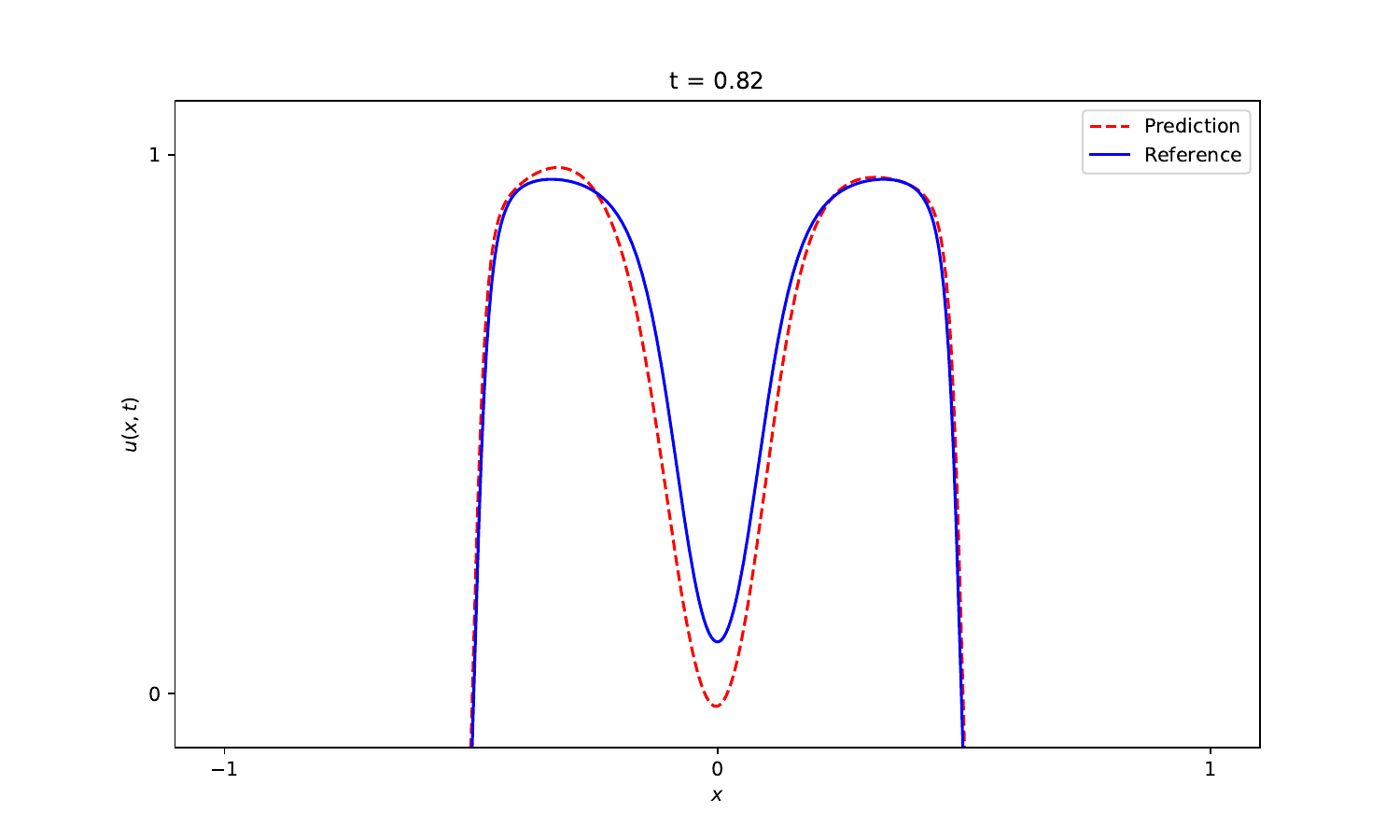}}
    \caption{Comparison of solutions at \( t=0.82 \) for the AC equation, with and without TL. Both subfigures display the reference solution alongside the PINN-predicted solution, using the lctanh AF in both cases.}
    \label{tl_comparison_ac_1}
\end{figure}
\vspace{-5em}
\begin{figure}[H]
    \centering
    \subfloat[tanh]{\includegraphics[width=0.55\textwidth]{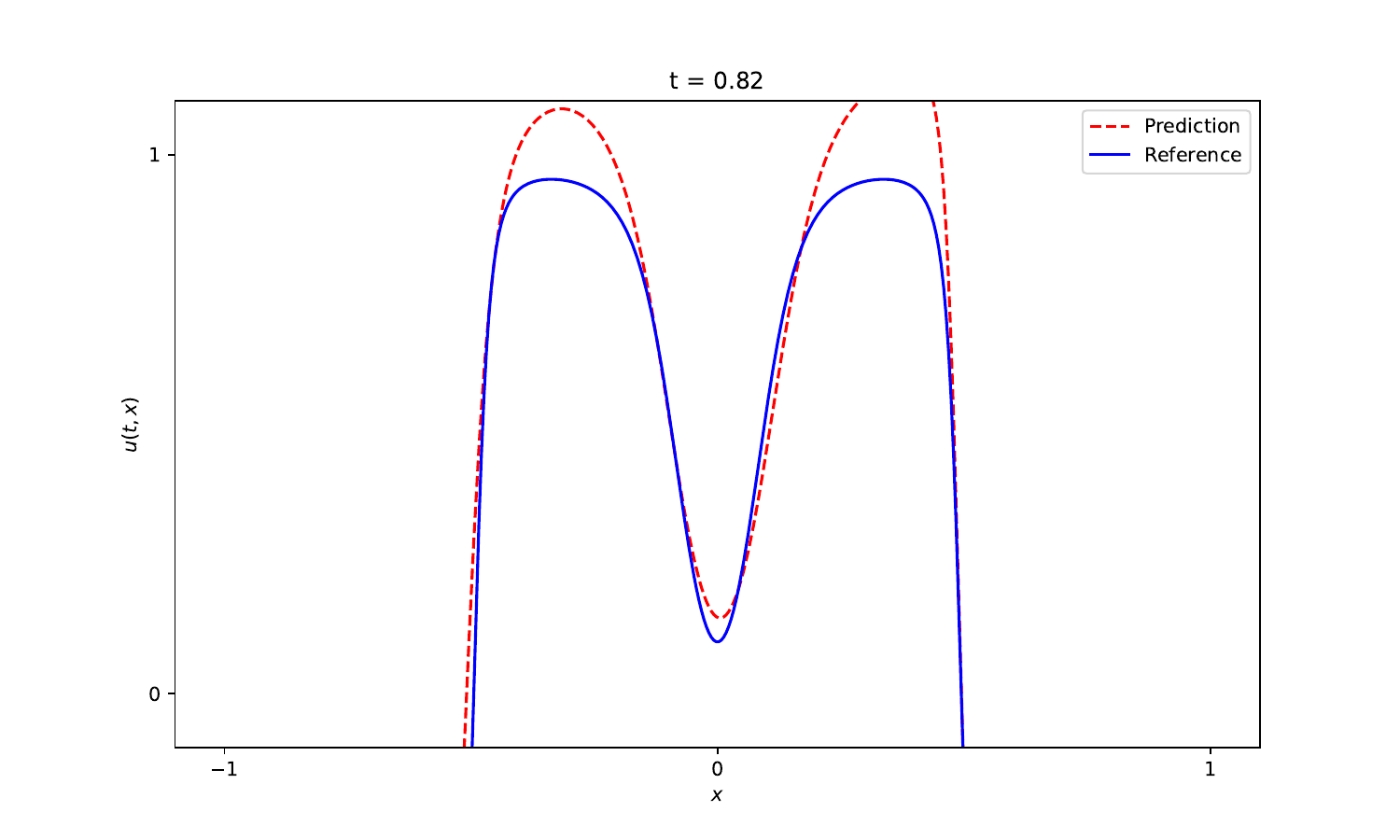}}
    \subfloat[lctanh]{\includegraphics[width=0.55\textwidth]{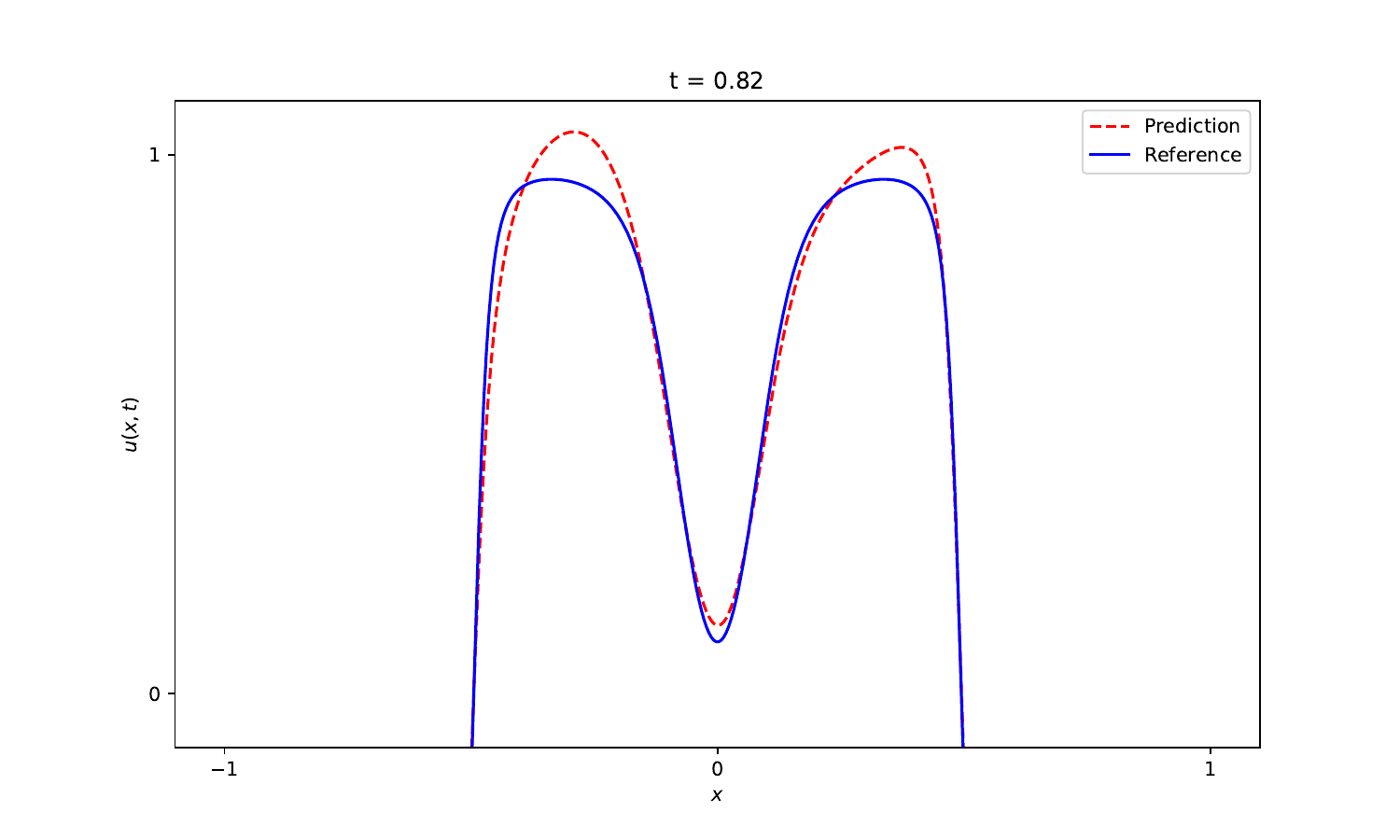}}
    \caption{Comparison of solutions at \( t=0.82 \) for the AC equation, generated with tanh and lctanh AF. Both subfigures display the reference solution alongside the PINN-predicted solution. No TL is used.}
    \label{tanh_comparison_ac_1}
\end{figure}

\begin{figure}[H]
    \centering
    \subfloat[Without TL]{\includegraphics[width=0.55\textwidth]{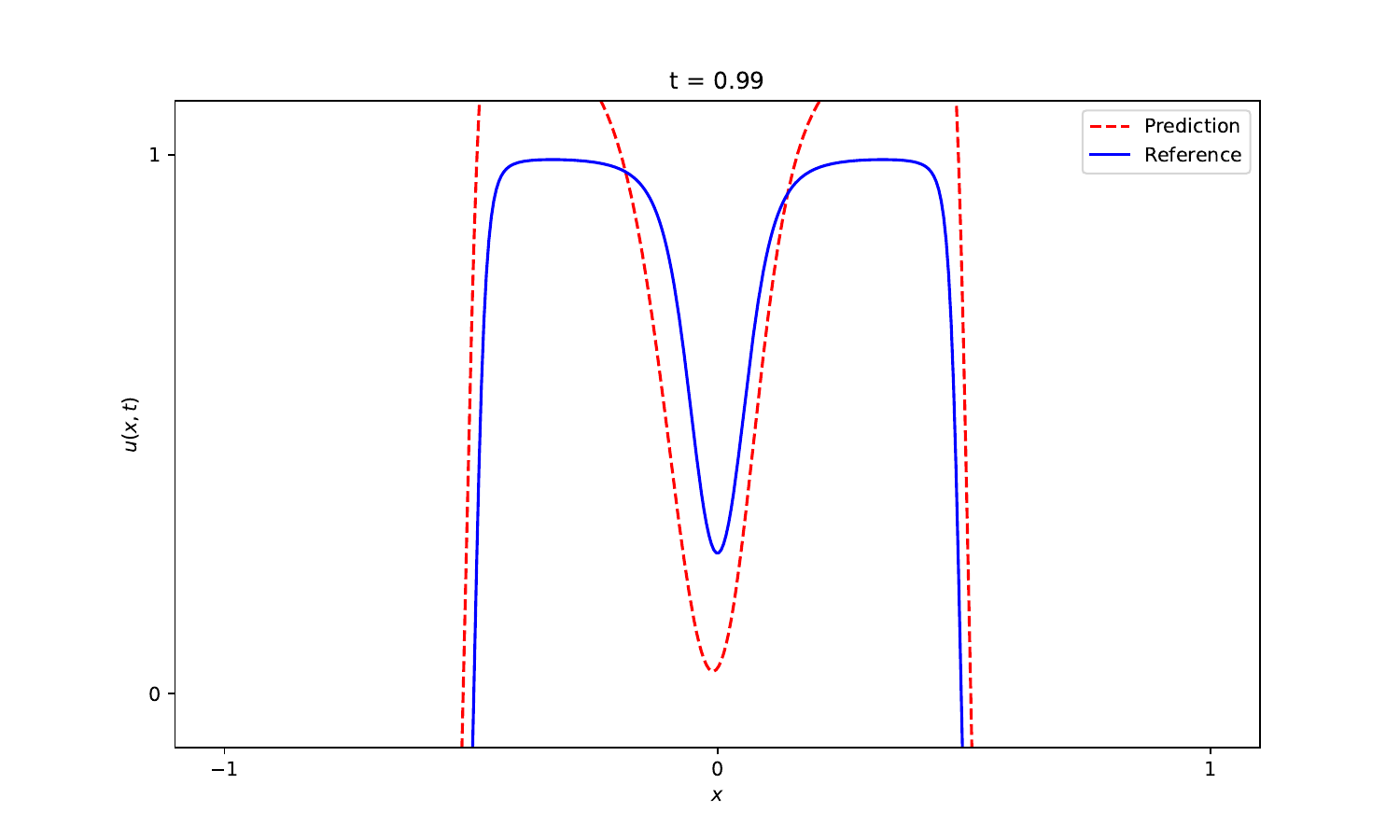}}
    \subfloat[With TL]{\includegraphics[width=0.55\textwidth]{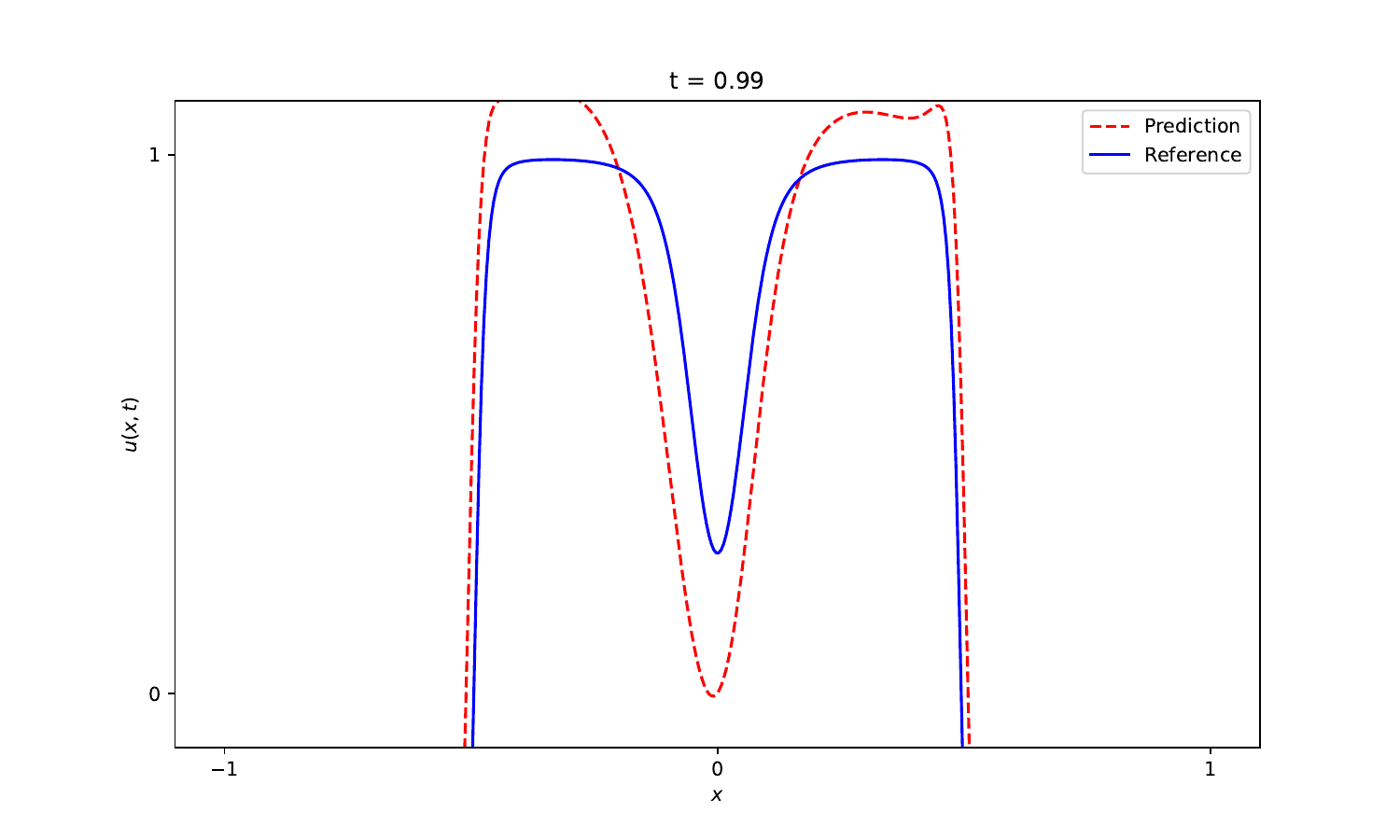}}
    \caption{Comparison of solutions at \( t=0.99 \) for the AC equation, with and without TL. Both subfigures display the reference solution alongside the PINN-predicted solution, using the lctanh AF in both cases.}
    \label{tl_comparison_ac_2}
\end{figure}

\begin{figure}[H]
    \centering
    \subfloat[tanh]{\includegraphics[width=0.55\textwidth]{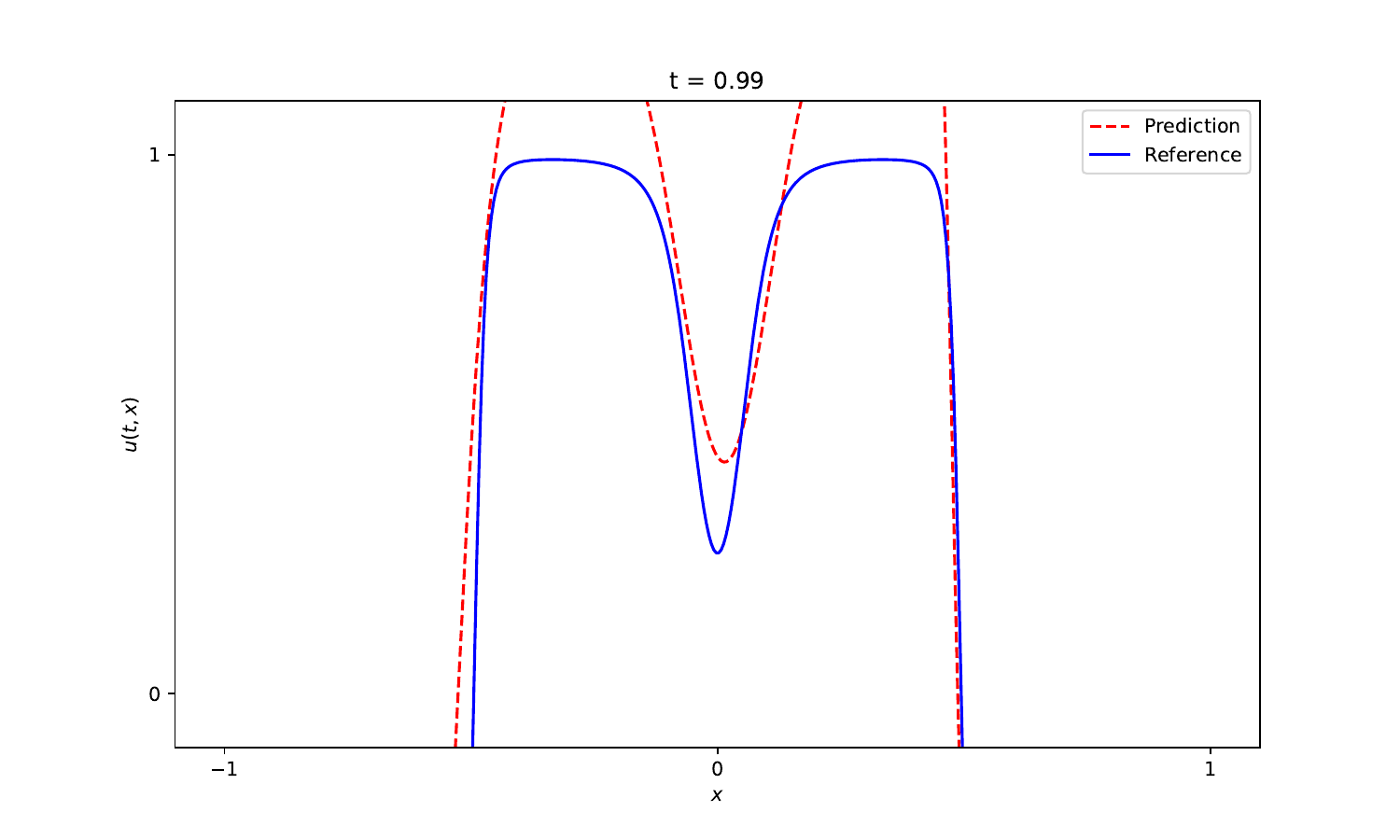}}
    \subfloat[lctanh]{\includegraphics[width=0.55\textwidth]{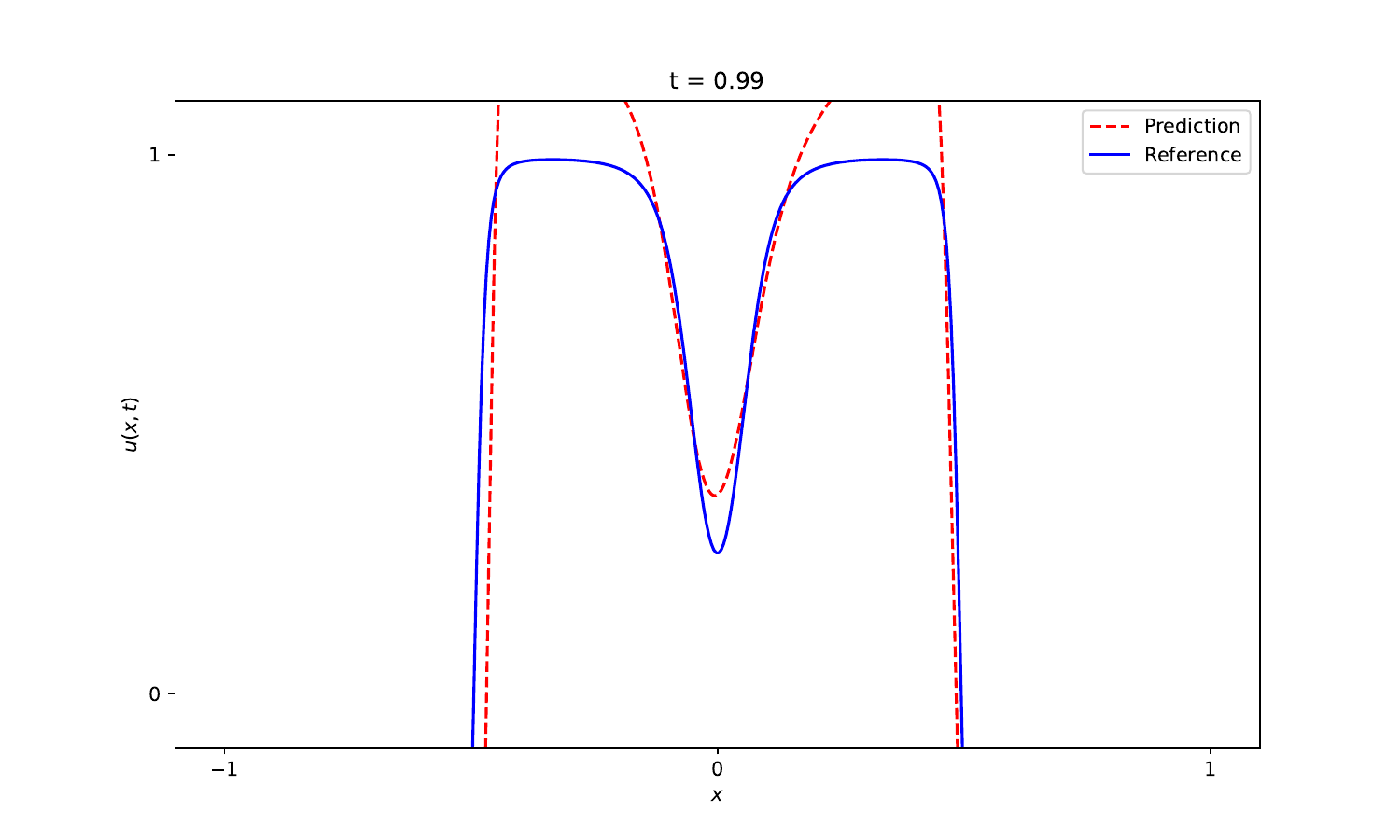}}
    \caption{Comparison of solutions at \( t=0.99 \) for the AC equation, generated with tanh and lctanh AF. Both subfigures display the reference solution alongside the PINN-predicted solution. No TL is used.}
    \label{tanh_comparison_ac_2}
\end{figure}

\subsection{KdV equation}

The KdV equation is given by
\begin{equation}
\begin{split}
&u_{t}+u u_{x}+0.0025 u_{x x x}=0, \text{ } x \in[-1,1], \text{ } t \in[0,1], \\
&u(0, x)=\cos (\pi x), u(t,-1)=u(t, 1), u_{x}(t,-1)=u_{x}(t, 1).\\
\end{split}
\end{equation}

\noindent We enforce the initial condition as a hard constraint by setting 

$$
u(t, x)=\cos (\pi x)+t v(t, x) 
$$

\noindent Then $u$ automatically satisfies the initial condition, and we only need to impose the PDE constraint and the boundary conditions on $v$. Then $v$ needs to satisfy
\begin{equation} \label{kdveqv}
\begin{split}
&v+tv_{t}+\left( \cos (\pi x)+t v \right)\left( -\pi \sin(\pi x)+t v_{x}\right)+0.0025 \left(\pi^3 \sin( \pi x) +tv_{xxx} \right)=0 \\
& v(t,-1)=v(t,1), v_{x}(t,-1)=v_{x}(t,1) . \\
\end{split}
\end{equation}

\begin{figure}[H]
    \centering
    \includegraphics[width=1.0\textwidth]{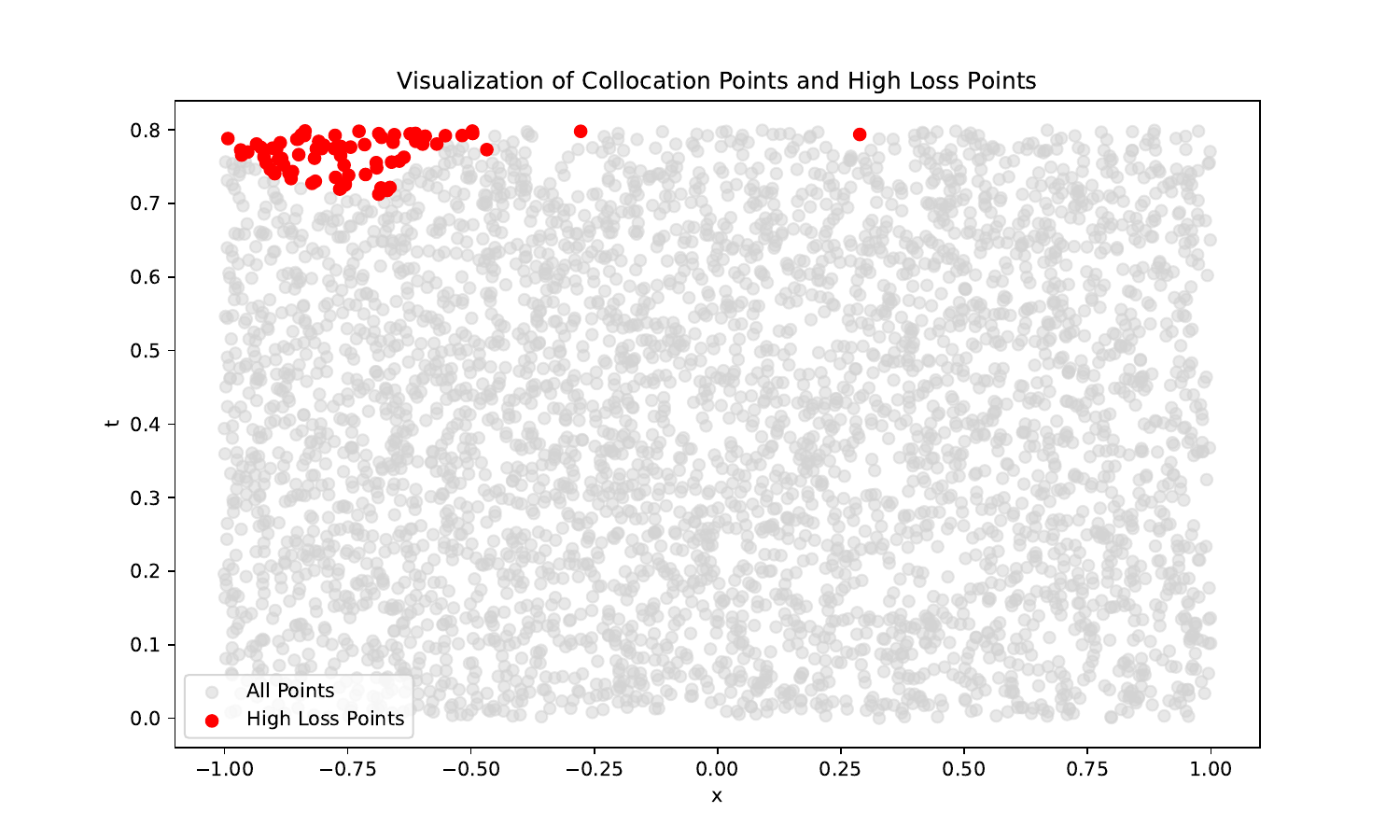}
    \caption{The collocation points with the highest PDE loss for the KdV equation are concentrated near the end of the temporal validation domain and around the regions of largest discrepancy in the spatial domain.}
    \label{coll_points_kdv}
\end{figure}
\vspace{-6em}
\begin{figure}[H]
    \centering
    \subfloat[Without TL]{\includegraphics[width=0.55\textwidth]{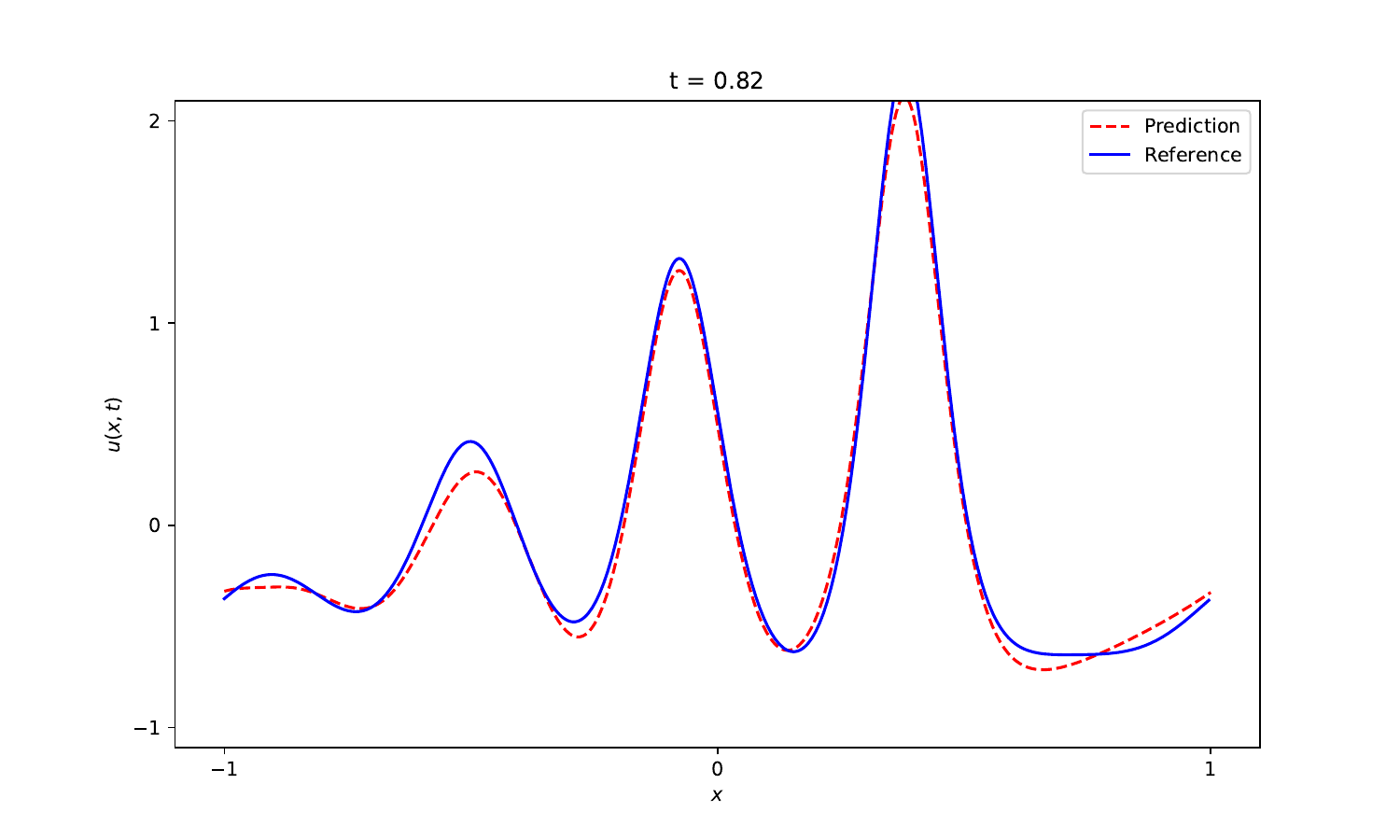}}
    \subfloat[With TL]{\includegraphics[width=0.55\textwidth]{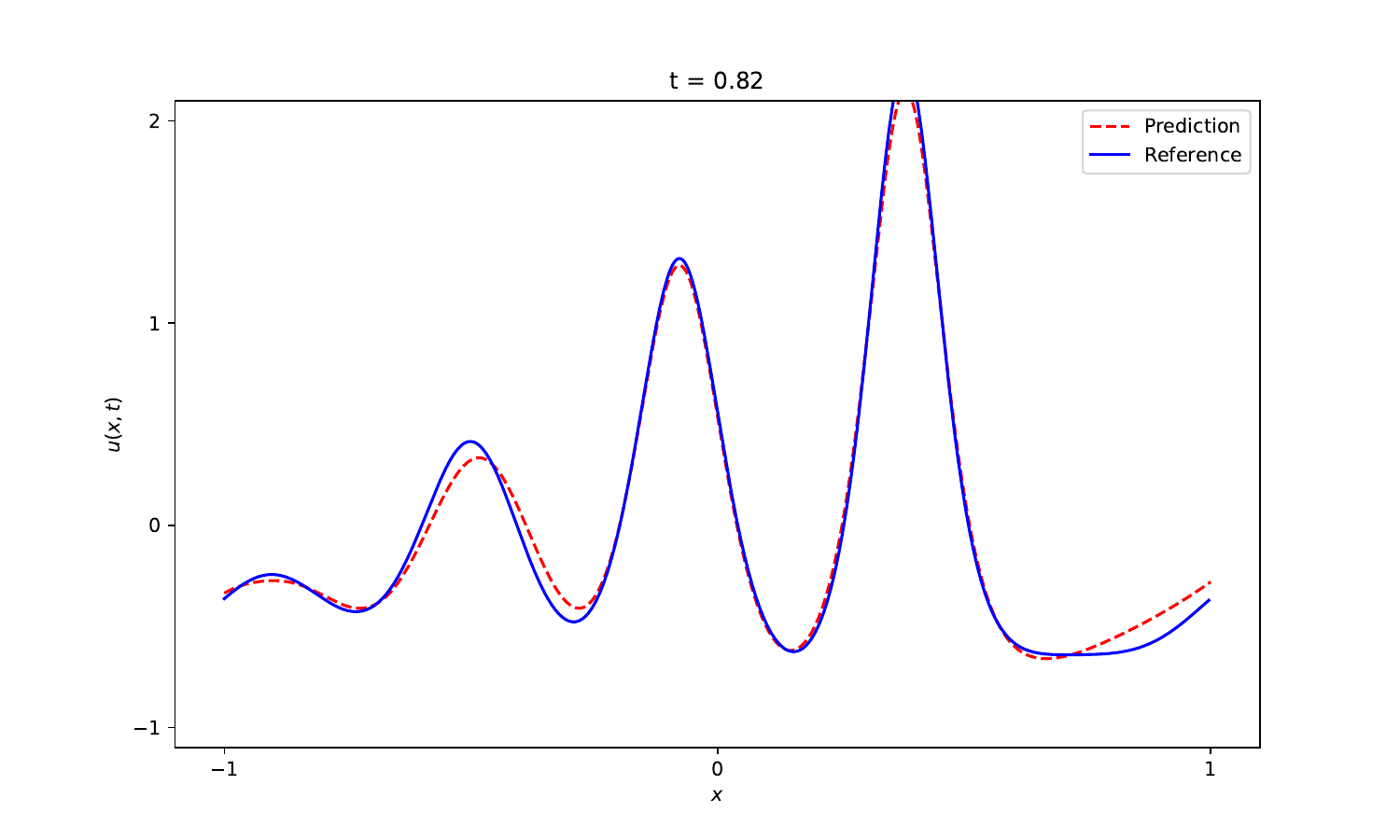}}
    \caption{Comparison of solutions at \( t=0.82 \) for the KdV equation, with and without TL. Both subfigures display the reference solution alongside the PINN-predicted solution, using the lc$(x + sin^2(x))$ AF in both cases.}
    \label{tl_comparison_kdv_1}
\end{figure}
\vspace{-5em}
\begin{figure}[H]
    \centering
    \subfloat[tanh]{\includegraphics[width=0.55\textwidth]{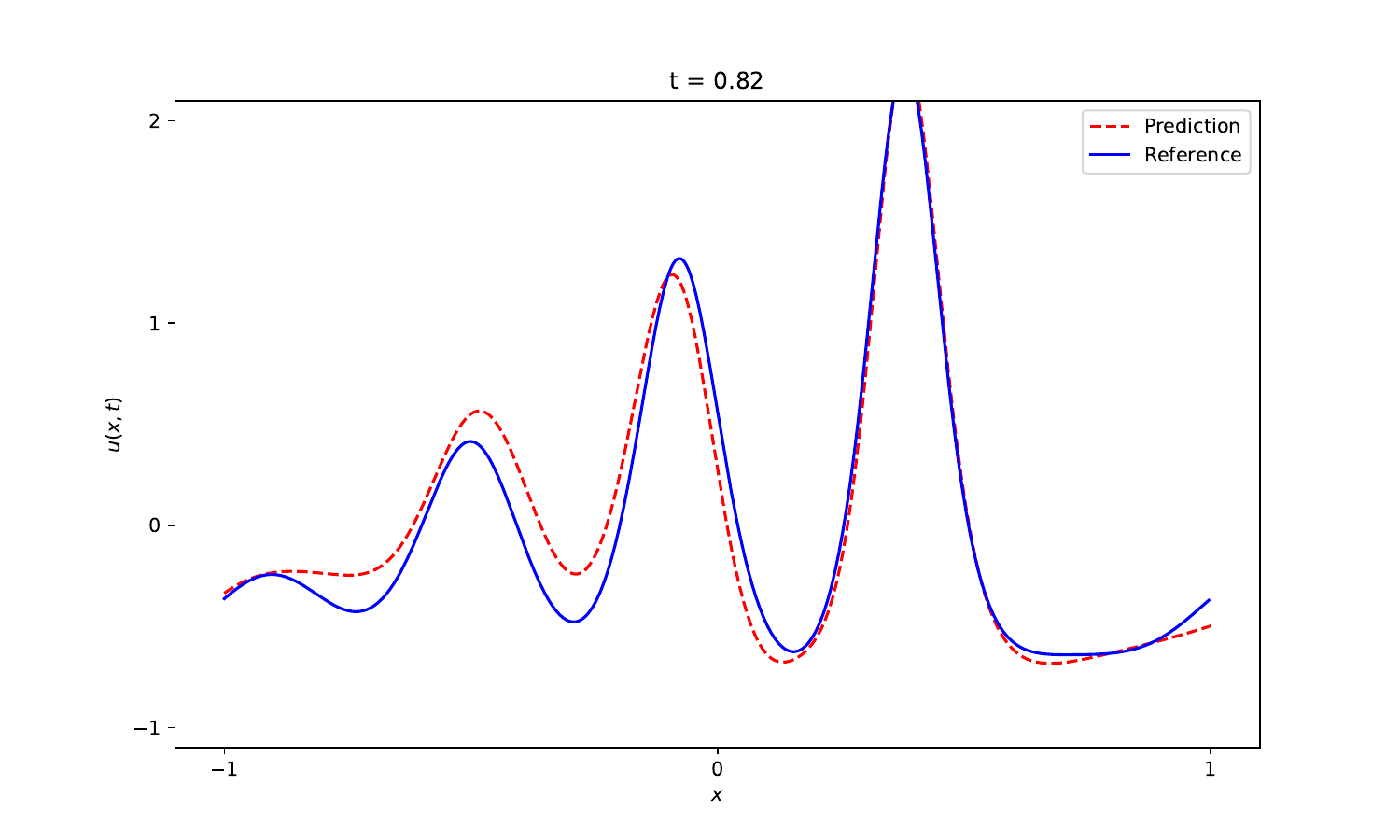}}
    \subfloat[lc $(x + sin^2(x))$]{\includegraphics[width=0.55\textwidth]{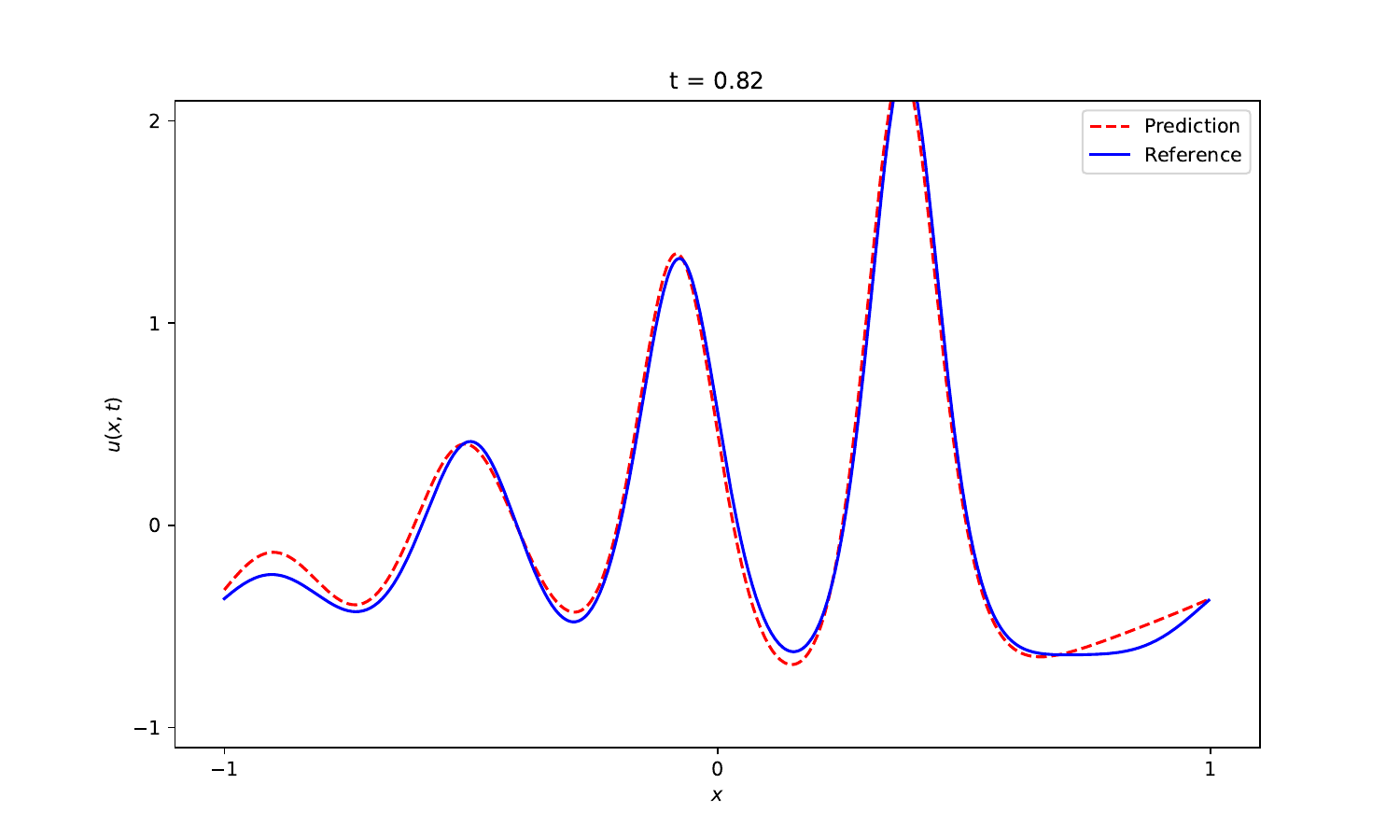}}
    \caption{Comparison of solutions at \( t=0.82 \) for the KdV equation, generated with tanh and lc $(x + sin^2(x))$ AF. Both subfigures display the reference solution alongside the PINN-predicted solution. No TL is used.}
    \label{tanh_comparison_kdv_1}
\end{figure}
\vspace{-5em}
\begin{figure}[H]
    \centering
    \subfloat[Without TL]{\includegraphics[width=0.55\textwidth]{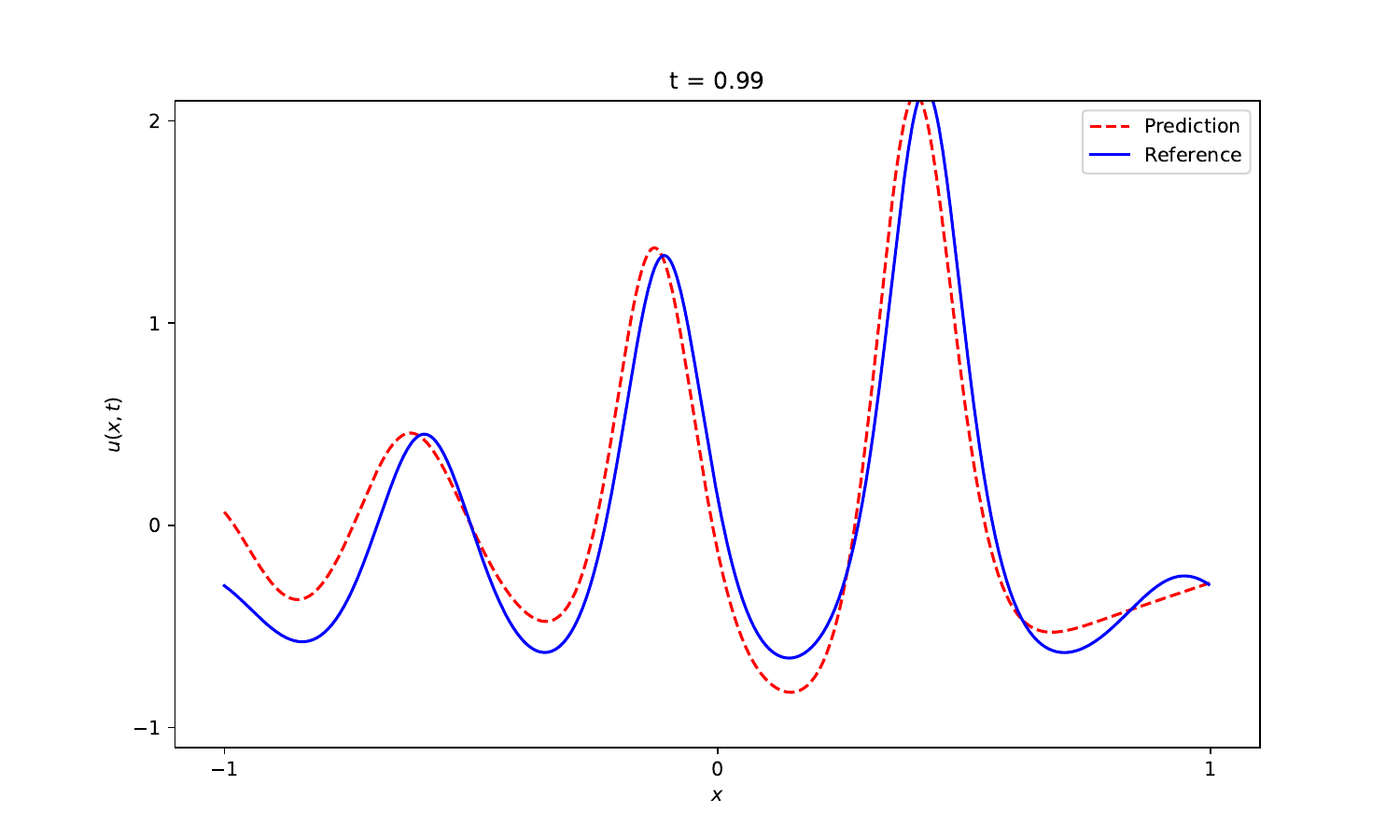}} 
    \subfloat[With TL]{\includegraphics[width=0.55\textwidth]{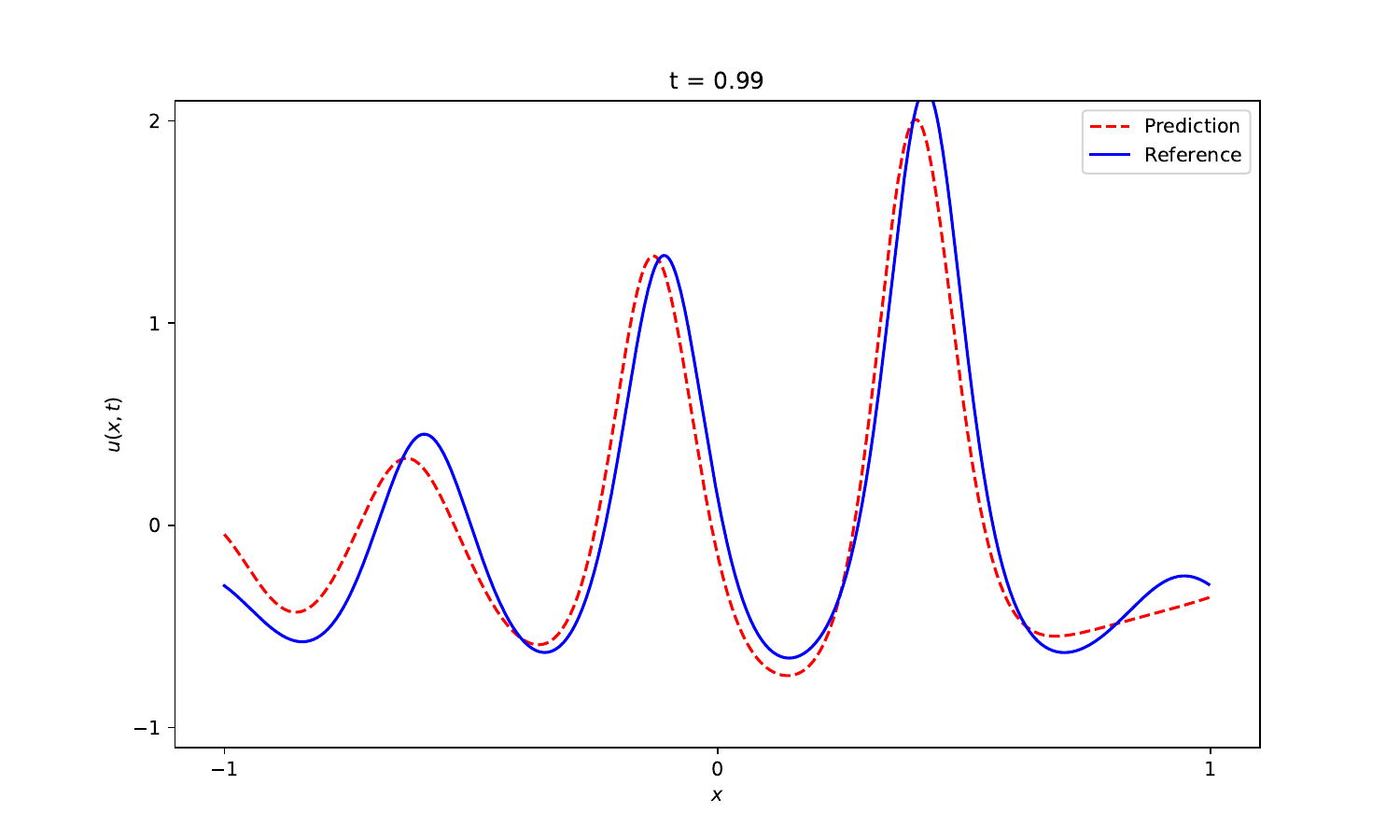}}
    \caption{Comparison of solutions at \( t=0.99 \) for the KdV equation, with and without TL. Both subfigures display the reference solution alongside the PINN-predicted solution, using the lc$(x + sin^2(x))$ AF in both cases.}
    \label{tl_comparison_kdv_2}
\end{figure}

\begin{figure}[H]
    \centering
    \subfloat[tanh]{\includegraphics[width=0.55\textwidth]{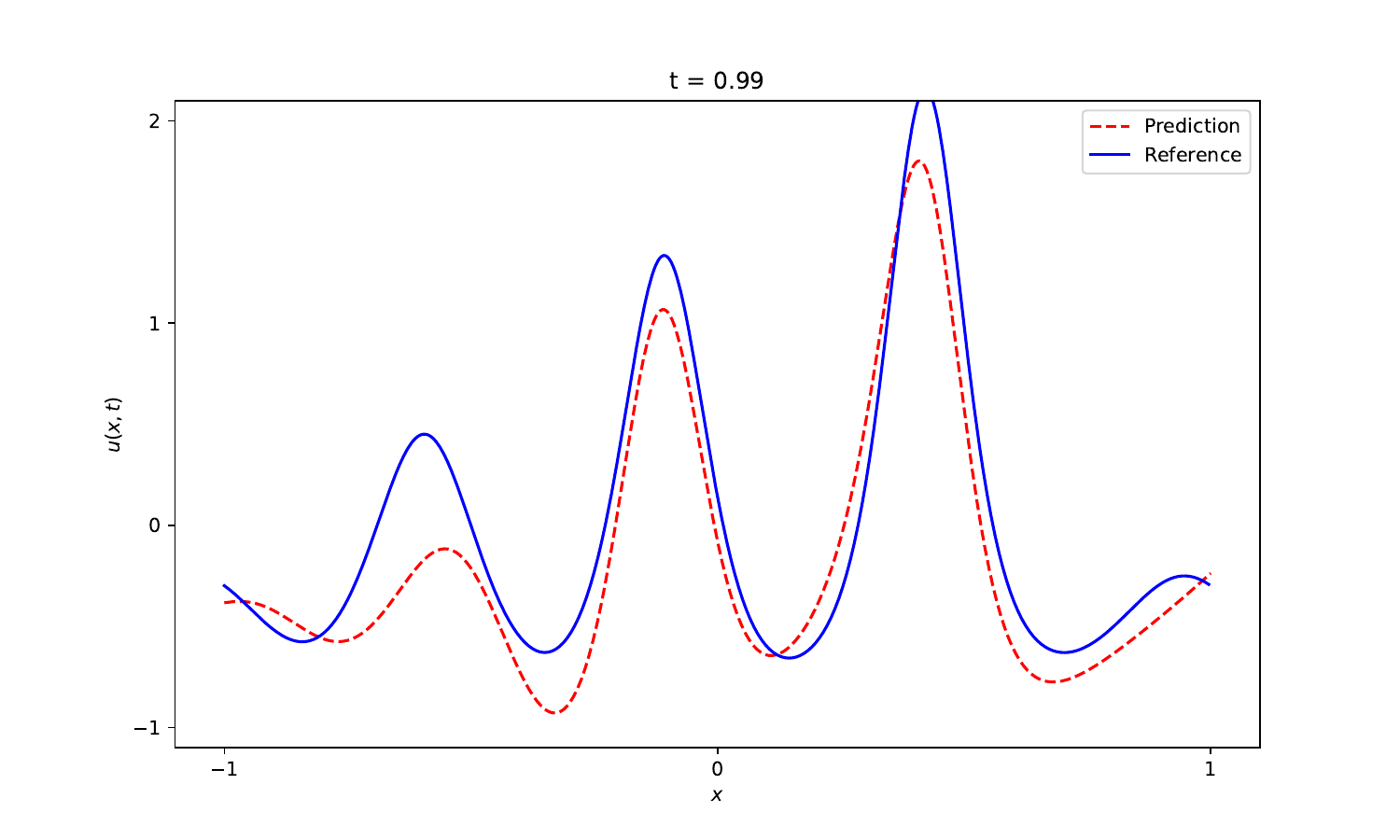}}
    \subfloat[lc $(x + sin^2(x))$]{\includegraphics[width=0.55\textwidth]{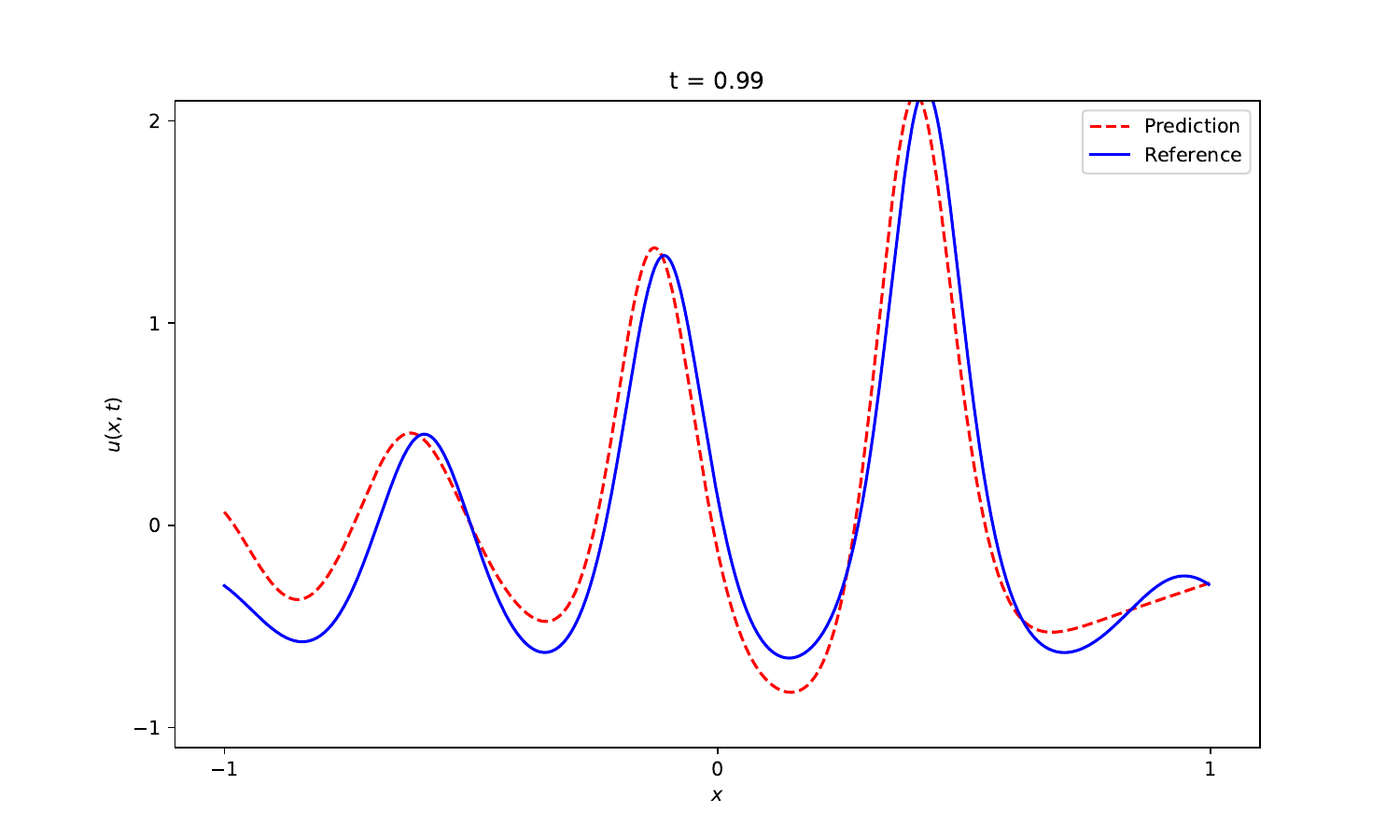}}
    \caption{Comparison of solutions at \( t=0.99 \) for the KdV equation, generated with tanh and lc $(x + sin^2(x))$ AF. Both subfigures display the reference solution alongside the PINN-predicted solution. No TL is used.}
    \label{tanh_comparison_kdv_2}
\end{figure}


\subsection{ Viscous Burgers' equation}
The viscous Burgers' equation is given by
\begin{equation}
\begin{split}
&u_{t}+u u_{x}-\frac{0.01}{\pi} u_{x x}=0, \text{ }x \in[-1,1], \text{ } t \in[0,1], \\
&u(0,x)=-\sin (\pi x), u(t,-1)=u(t, 1)=0.\\
\end{split}
\end{equation}

\noindent We enforce the initial and boundary conditions as hard constraints by setting 
$$
u(t,x)=-\sin (\pi x) + t(1-x^2)v(t,x)
$$

\noindent Then u automatically satisfies the initial and boundary conditions, and we only need to impose
the PDE constraint on $v$. The equation for $v$ is given by
\small
\begin{equation} \label{burgerseqv}
\begin{split}
& (1-x^2)v +t(1-x^2)v_{t} +\left( -\sin(\pi x)+t(1-x^{2})v \right) \left(-\pi \cos( \pi x) +t(-2x)v+t(1-x^{2})v_{x}\right)\\
&-\frac{0.01}{\pi} \left(\pi^2 \sin(\pi x) -2tv-4xtv_{x}+t(1-x^{2})v_{xx}\right)=0 \\
\end{split}
\end{equation}
\normalsize

\begin{figure}[H]
    \centering
    \includegraphics[width=1.0\textwidth]{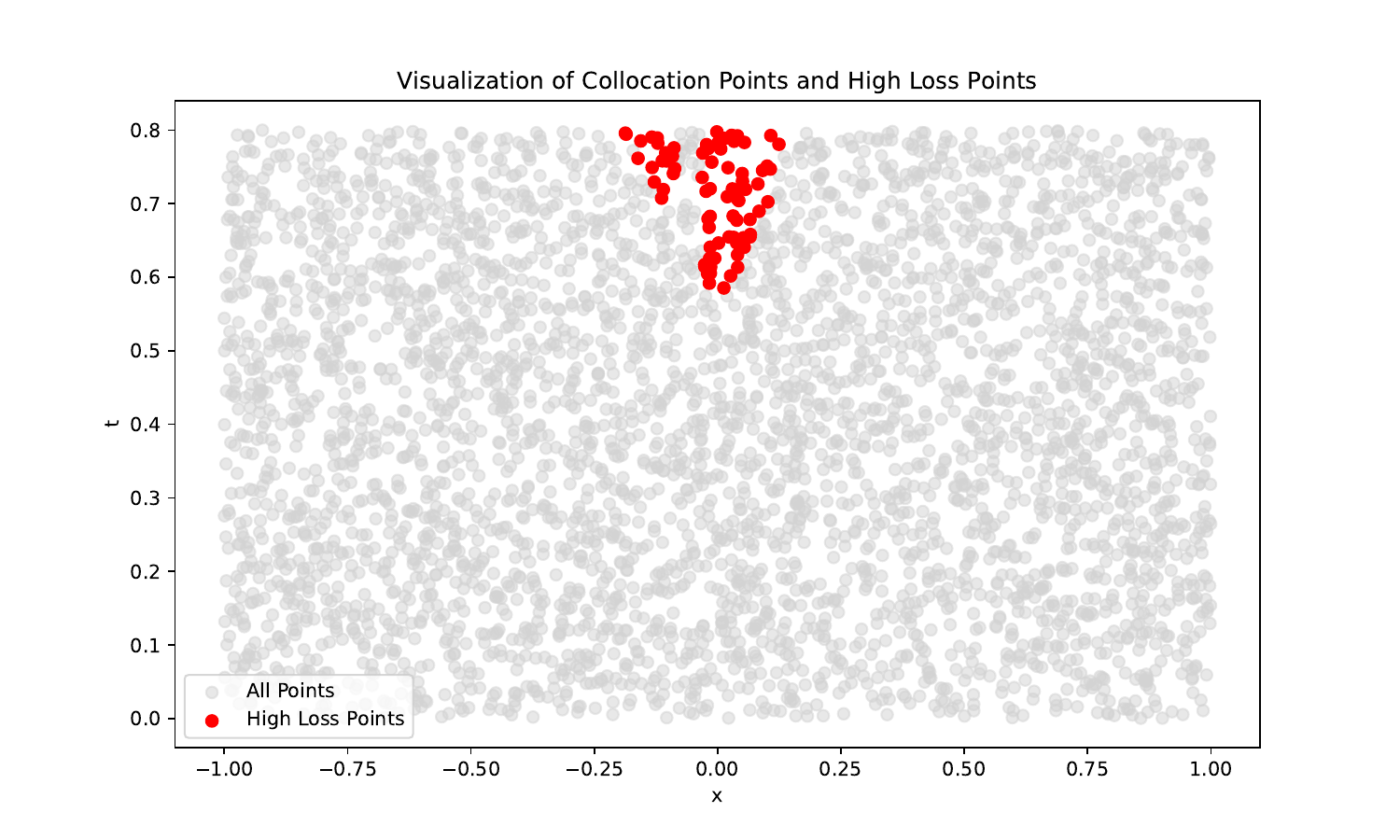}
    \caption{The collocation points with the highest PDE loss for the Burgers' equation are concentrated near the end of the temporal validation domain and around the regions of largest discrepancy in the spatial domain.}
    \label{coll_points_burgers}
\end{figure}

\begin{figure}[H]
    \centering
    \subfloat[Without TL]{\includegraphics[width=0.55\textwidth]{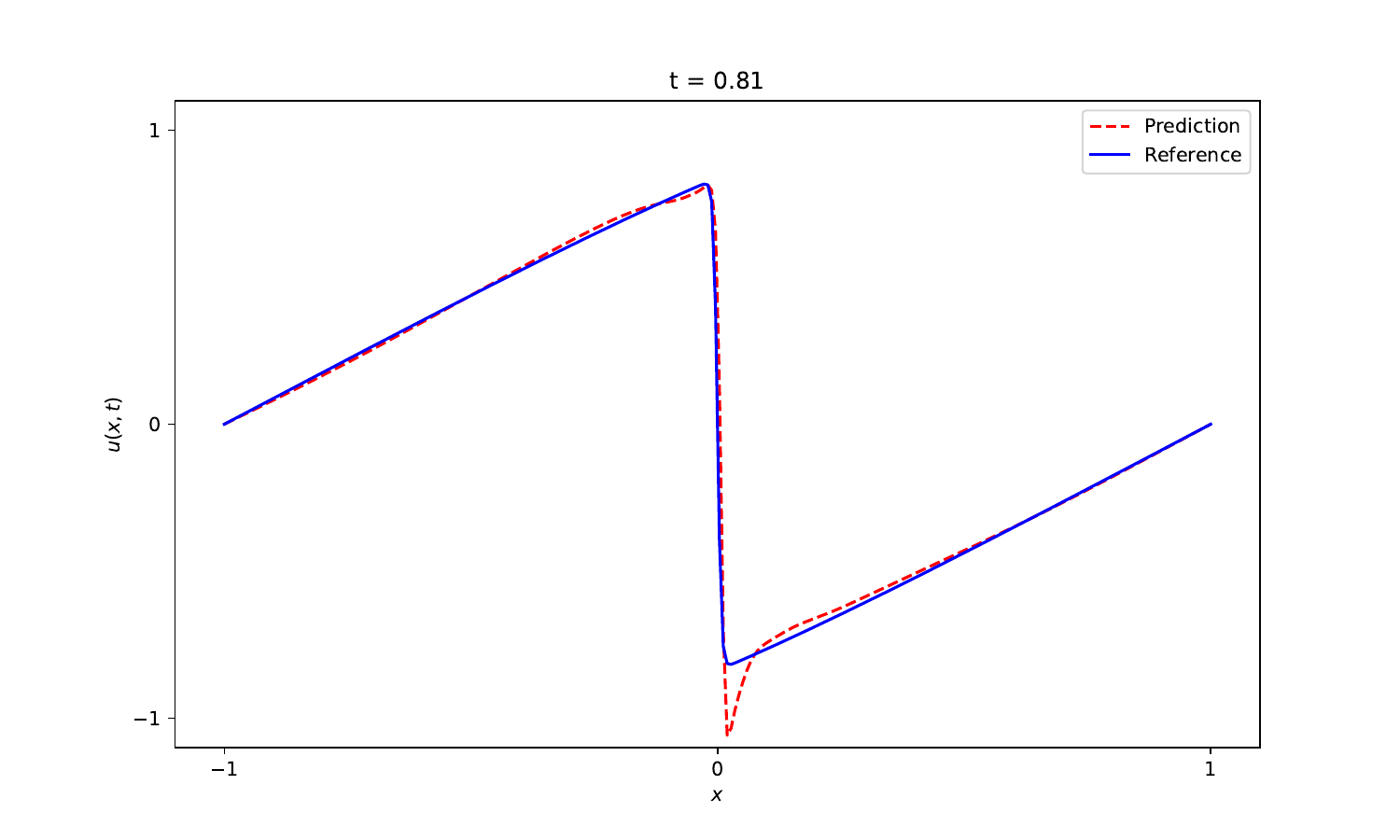}} 
    \subfloat[With TL]{\includegraphics[width=0.55\textwidth]{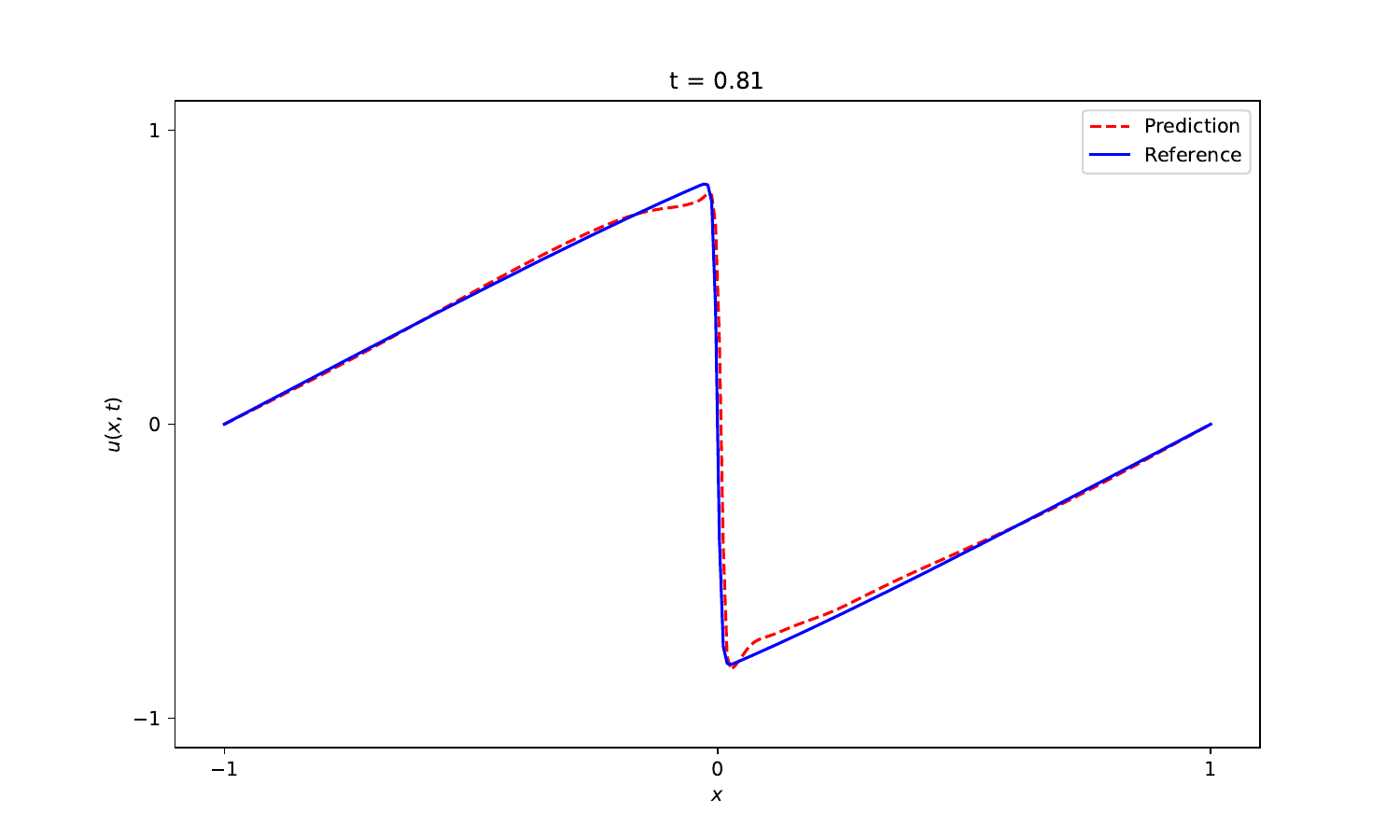}}
    \caption{Comparison of solutions at \( t=0.81 \) for the Burgers' equation, with and without TL. Both subfigures display the reference solution alongside the PINN-predicted solution, using the lctanh AF in both cases.}
    \label{tl_comparison_burgers_1}
\end{figure}
\vspace{-5em}
\begin{figure}[H]
    \centering
    \subfloat[tanh]{\includegraphics[width=0.55\textwidth]{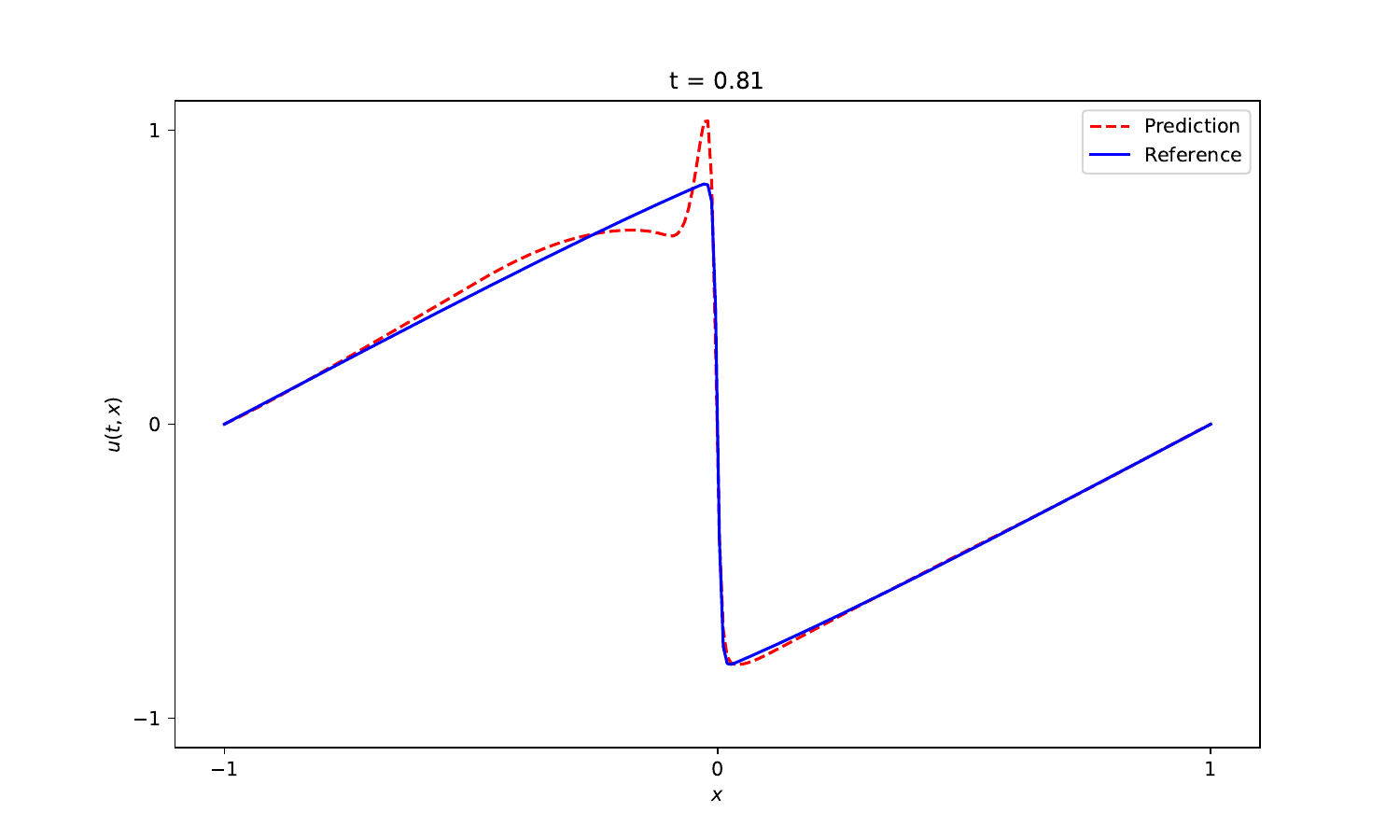}} 
    \subfloat[lctanh]{\includegraphics[width=0.55\textwidth]{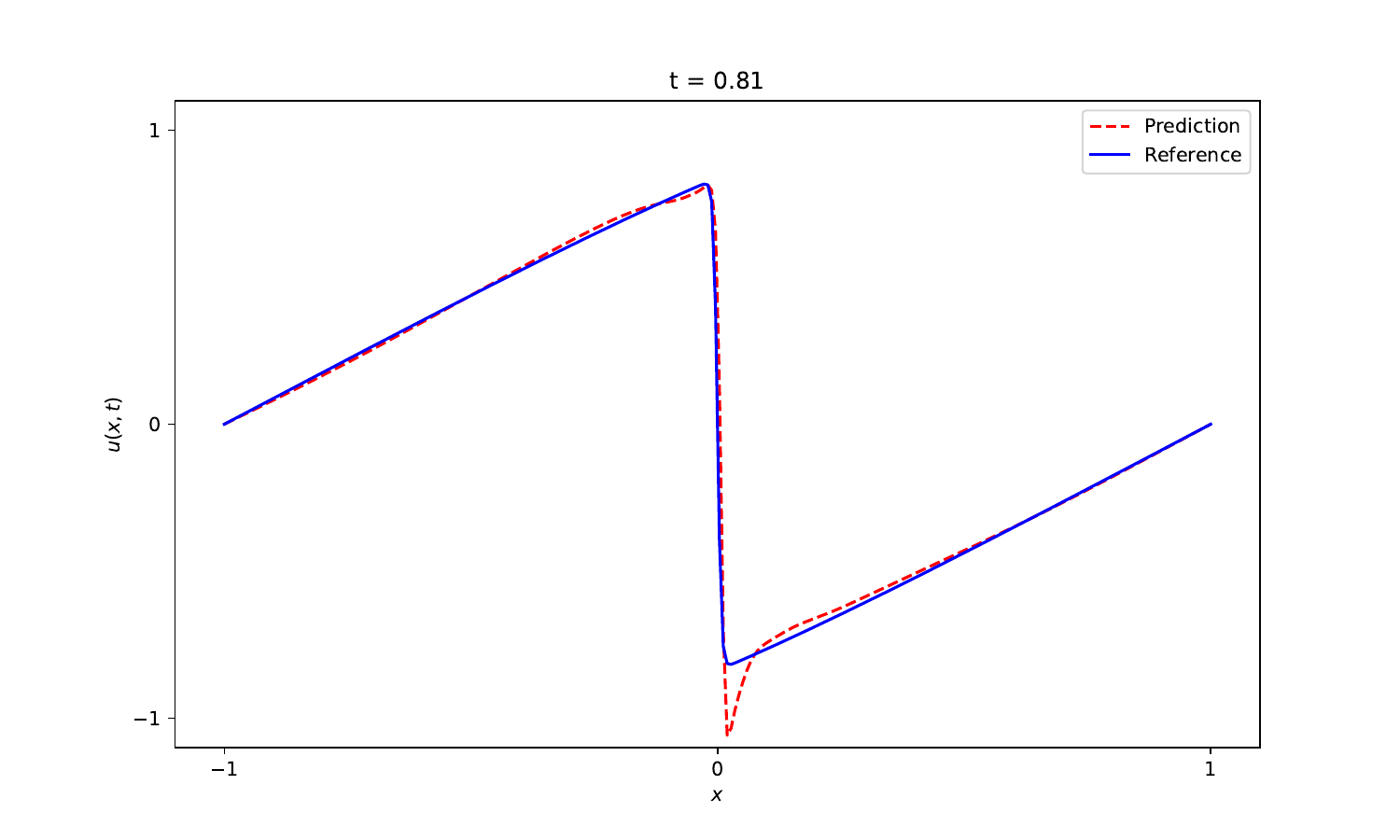}}
    \caption{Comparison of solutions at \( t=0.81 \) for the Burgers equation, generated with tanh and lctanh AF. Both subfigures display the reference solution alongside the PINN-predicted solution. No TL is used.}
    \label{tanh_comparison_burgers_1}
\end{figure}

\begin{figure}[H]
    \centering
    \subfloat[Without TL]{\includegraphics[width=0.55\textwidth]{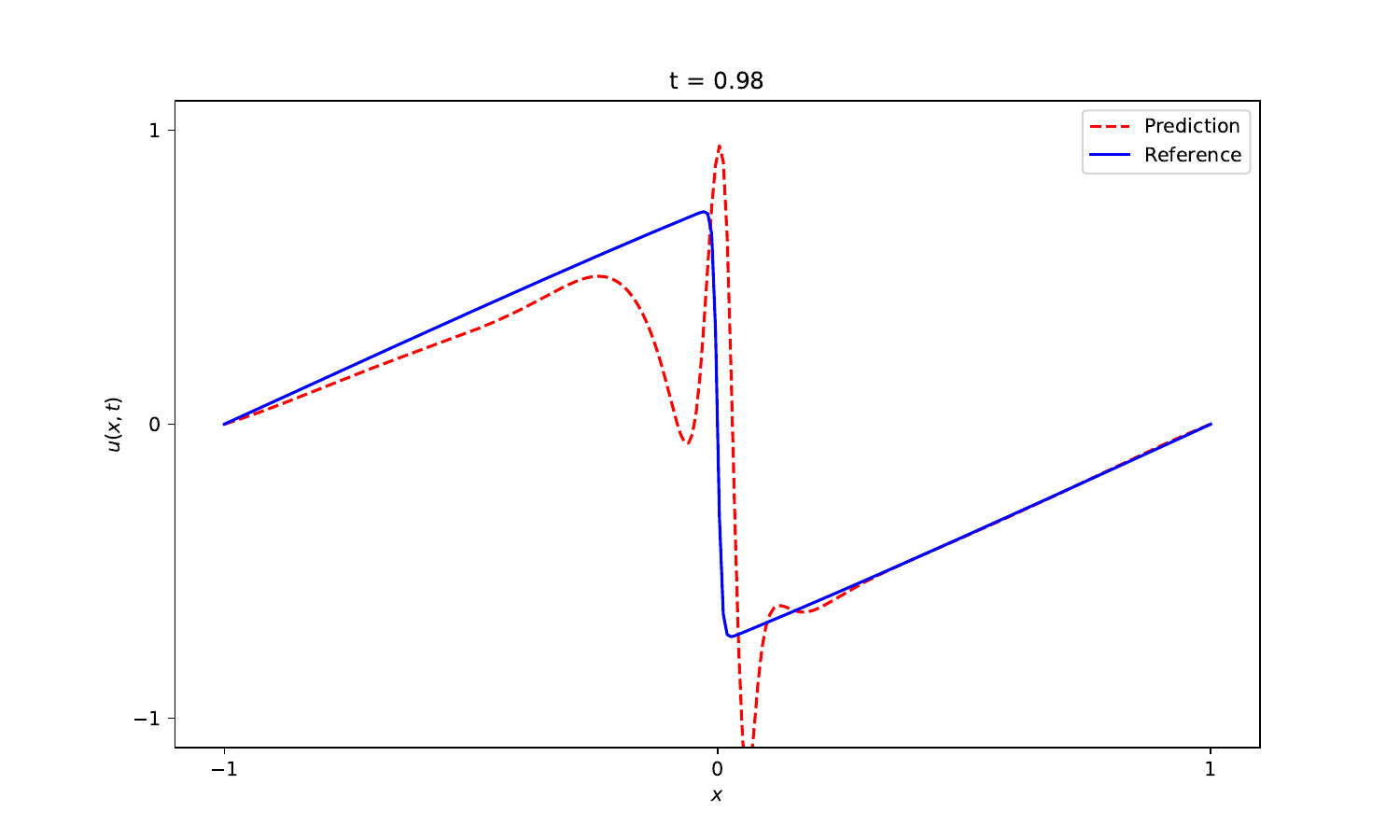}}
    \subfloat[With TL]{\includegraphics[width=0.55\textwidth]{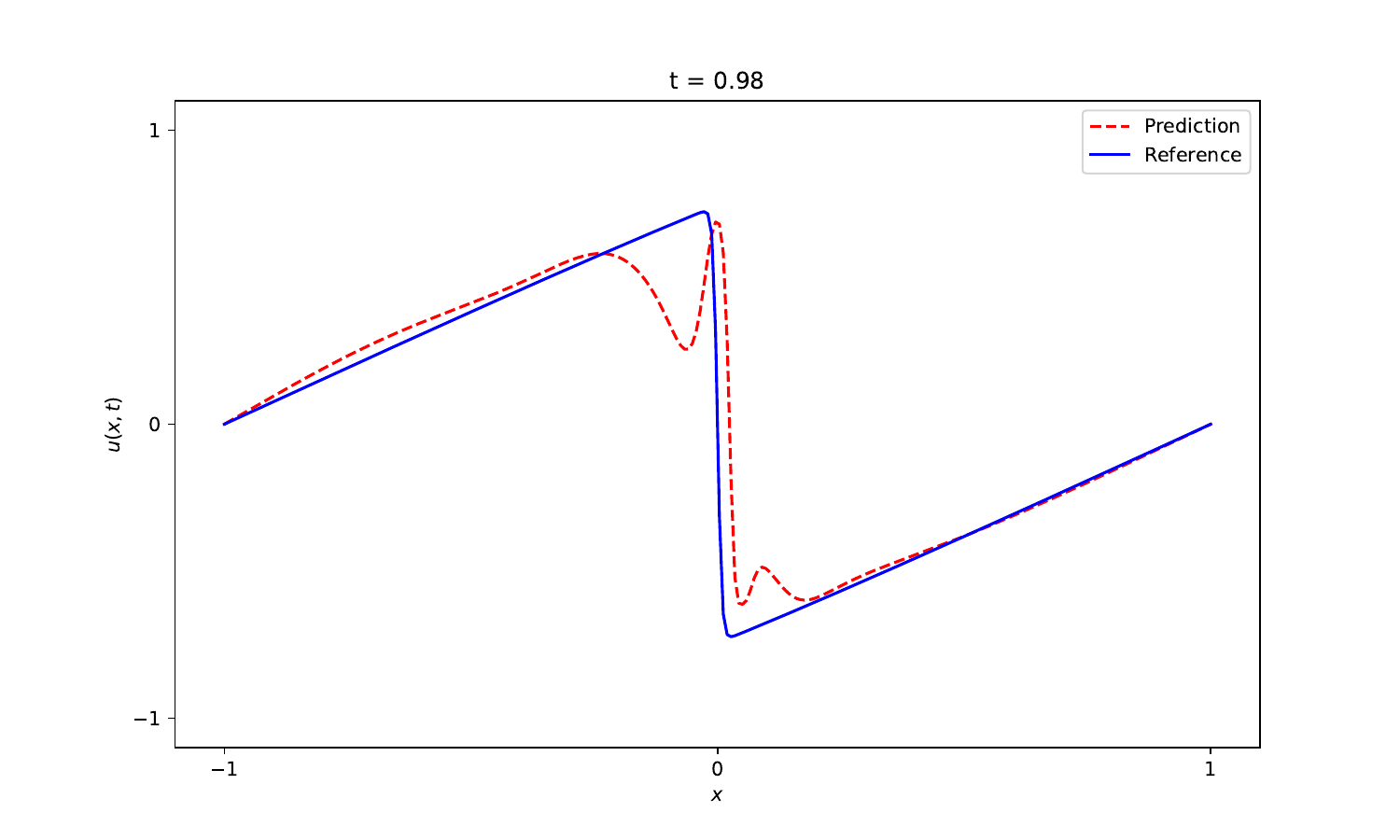}}
    \caption{Comparison of solutions at \( t=0.98 \) for the Burgers' equation, with and without TL. Both subfigures display the reference solution alongside the PINN-predicted solution, using the lctanh AF in both cases.}
    \label{tl_comparison_burgers_2}
\end{figure}

\begin{figure}[H]
    \centering
    \subfloat[tanh]{\includegraphics[width=0.55\textwidth]{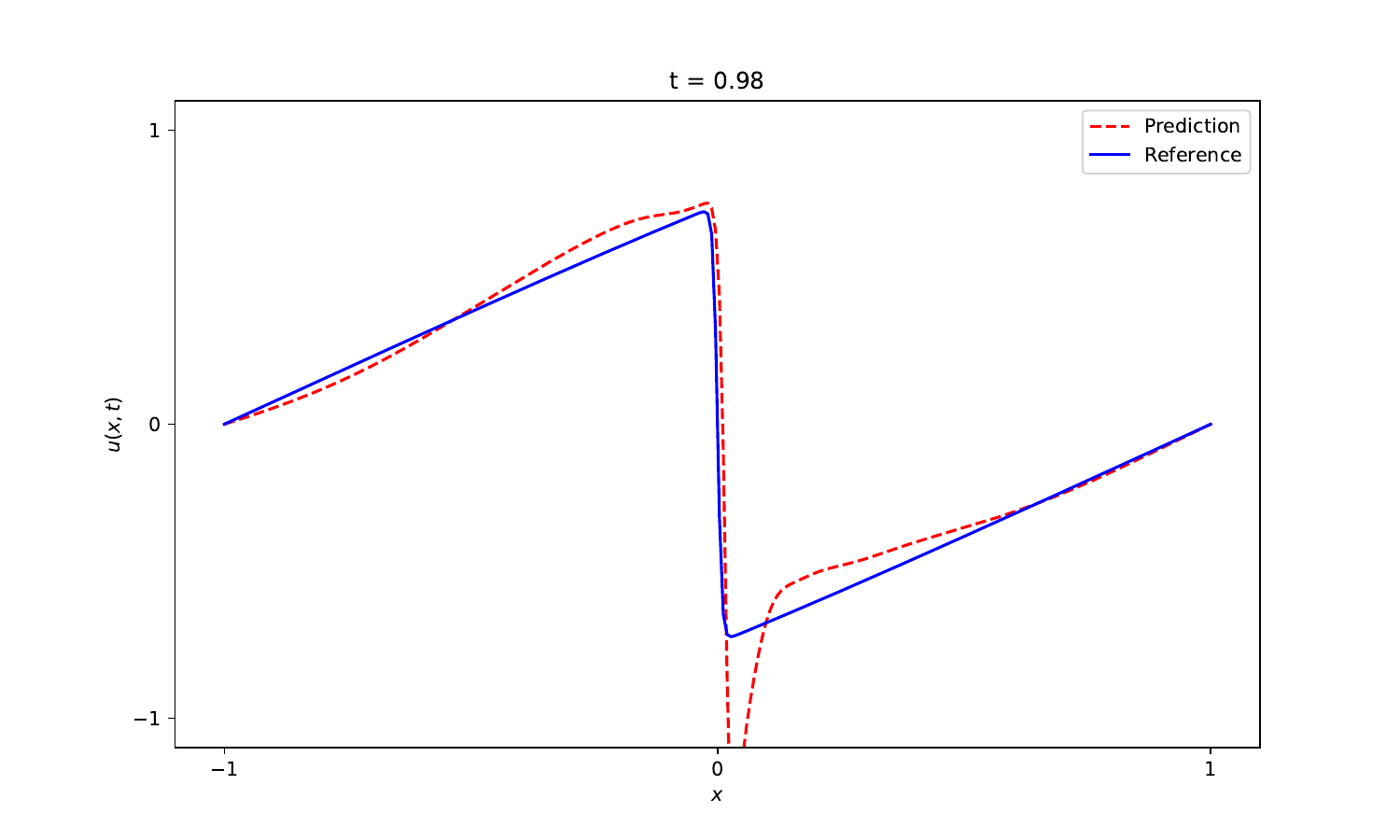}} 
    \subfloat[lctanh]{\includegraphics[width=0.55\textwidth]{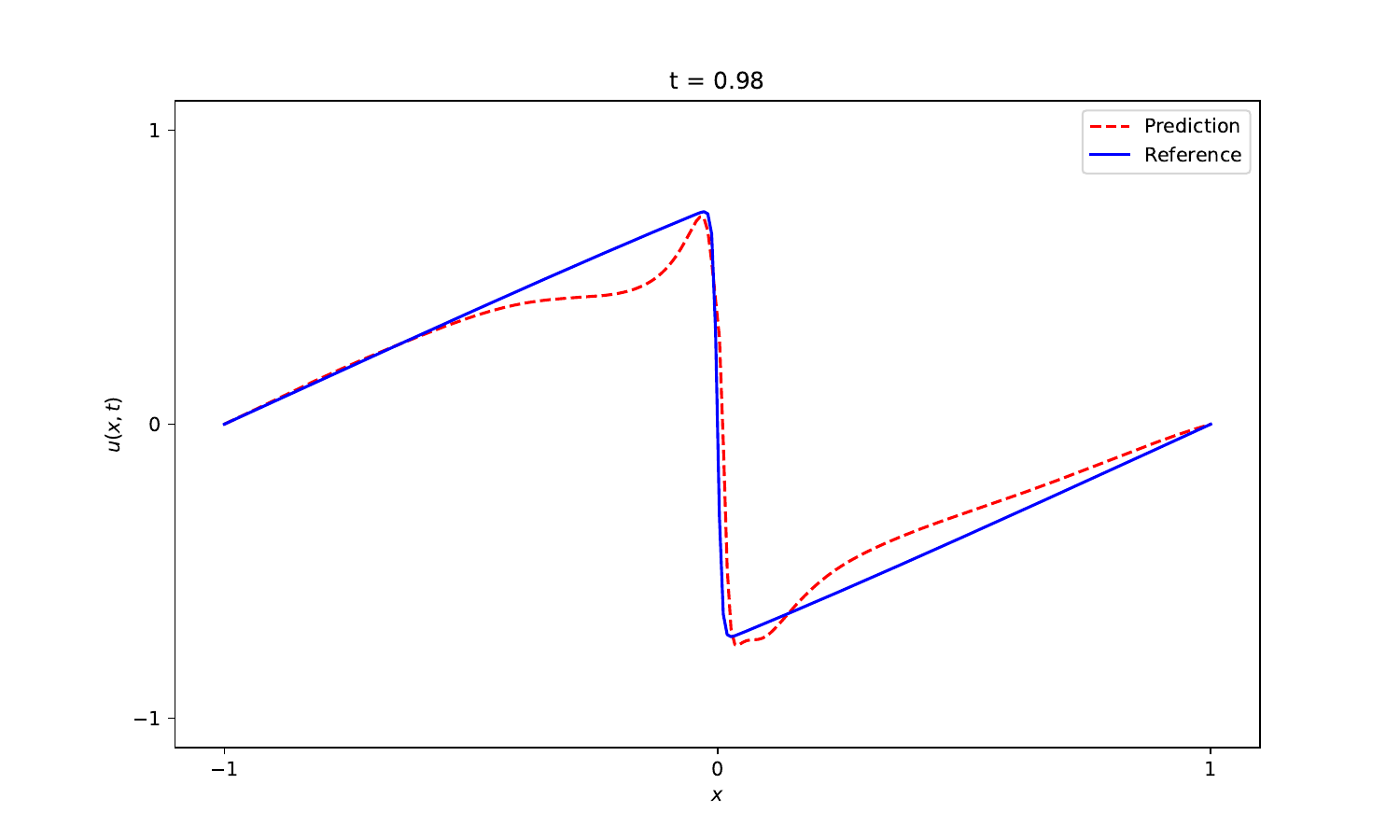}}
    \caption{Comparison of solutions at \( t=0.98 \) for the Burgers equation, generated with tanh and lctanh AF. Both subfigures display the reference solution alongside the PINN-predicted solution. No TL is used.}
    \label{tanh_comparison_burgers_2}
\end{figure}

\section{Computational efficiency}

We report the training times of our methods applied to the AC equation in Table S\ref{trainingtimes} below. 

\begin{table}[ht] 
\begin{center}
\begin{tabular}{|l|l|}
\hline
Method        & Training Time (min) \\ \hline
tanh w/o TL   & 51.6                \\ \hline
lctanh w/o TL & 58.3                \\ \hline
tanh w/ TL    & 53.2                \\ \hline
lctanh w/TL   & 60.7                \\ \hline
\end{tabular}
\end{center}
\caption{Training time comparison for the vanilla PINN (tanh) and our improved AF PINN (lctanh), with and without TL, for the AC equation. The results show that modifying the activation function or incorporating TL has only a minimal effect on the training time.}
\label{trainingtimes}
\end{table}

\section{Methodological rationale: Final-layer fine-tuning and activation function selection}
We introduce an adaptive AF only on the final layer, because, based on our findings (omitted), altering the AF throughout the network can negatively affect convergence and lead to training instability.

\begin{figure}[ht] 
    \centering
    \includegraphics[width=0.7\textwidth]{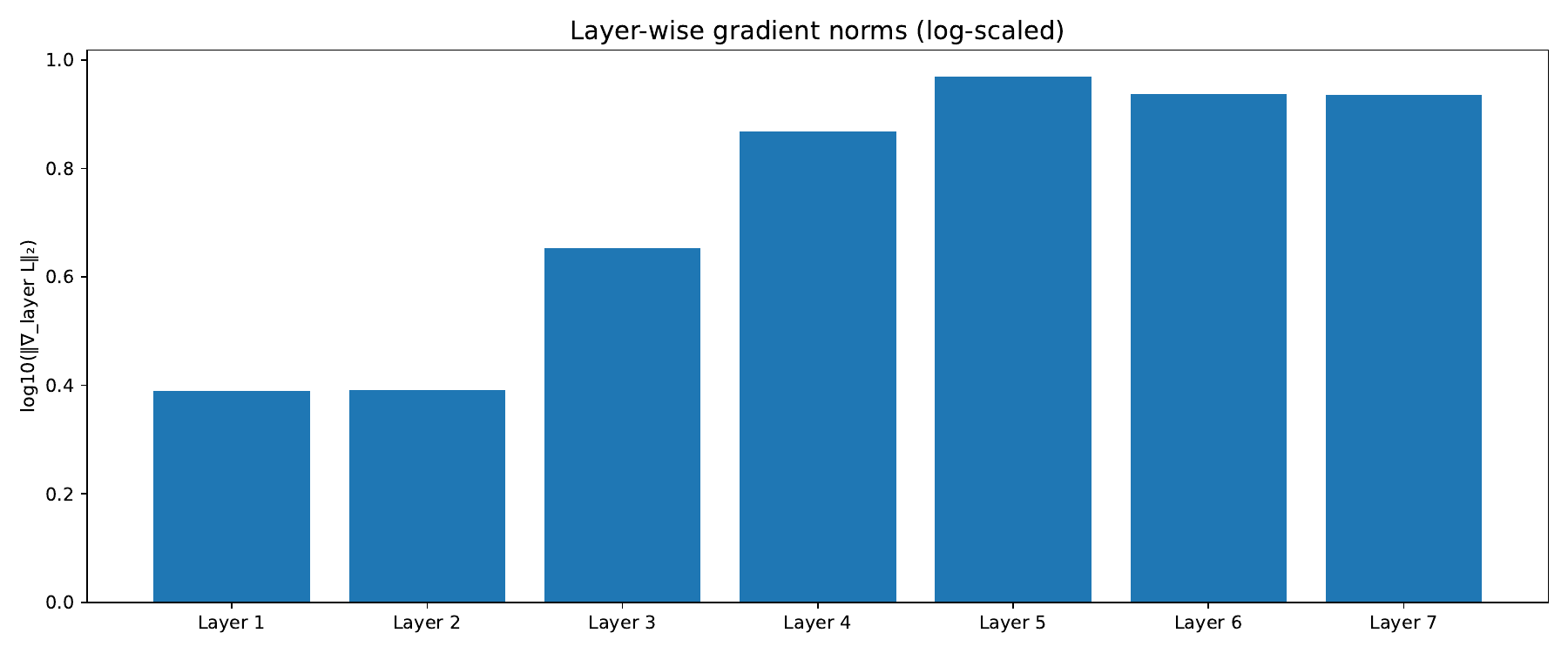}
    \caption{Layer-wise gradient magnitudes (log-scaled $L_2$ norms) of the weight and bias terms, illustrating the contribution of each layer. The highest magnitudes are observed in the final hidden layers, suggesting their dominant role in the model's performance. Results are computed for the AC equation with the tanh AF.}
    \label{grad_norms}
\end{figure}

To further justify fine-tuning only the final layer, we performed a layer-wise analysis on the AC equation using the tanh AF. We first computed the $L_2$ norms of the weight and bias gradients, then took the logarithm of these norms and summed them together. We then visualized them in Figure S\ref{grad_norms}. Our results reveal that the final three layers exhibit the largest values, indicating that these layers contribute most significantly to task-specific adaptation. This aligns with established transfer learning practices \cite{desai_one-shot_2021}, where fine-tuning is typically performed on the final layers while earlier layers are frozen.

Regarding the AF selection, we experimented comparing ABU-PINN and lctanh for the AC equation. Our results show that, on average, lctanh achieved a training loss of \texttt{1e-5} after 330 epochs, compared to 360 epochs for ABU-PINN, suggesting faster optimization. Additionally, lctanh began experiencing an increasing validation loss around epoch 720, while ABU-PINN did so around epoch 680, indicating that lctanh is less prone to overfitting. These findings provide evidence that the linear combination of tanh AFs can offer improved optimization and reduced overfitting compared to ABU-PINN, in addition to the PDE-specific reasons mentioned in the main text.

\end{document}